\documentclass{article}

\usepackage[final]{nips_2016_arxiv}

\usepackage[utf8]{inputenc}  
\usepackage[T1]{fontenc}     
\usepackage[dvips]{hyperref} 
\usepackage{url}             
\usepackage{booktabs}        
\usepackage{amsfonts}        
\usepackage{microtype}       
\usepackage{nicefrac}        

\setcitestyle{authoryear,open={(},close={)}}

\usepackage{graphicx}
\graphicspath{{figs/}}
\renewcommand{\figurename}{Fig.}
\usepackage[font=small]{caption}

\usepackage{float}
\usepackage{algorithm} 
\usepackage[noend]{algpseudocode}
\algrenewcommand\algorithmicfunction{\textbf{def}}
\algrenewcommand\algorithmicrequire{\textbf{Input:}}
\algrenewcommand\algorithmicensure{\textbf{Output:}}

\usepackage{amsmath,amssymb}
\usepackage{xcolor}
\usepackage[utf8]{inputenc}
\usepackage{listings}
\usepackage{upquote}
\usepackage{float}

\usepackage{algorithm} 
\usepackage[noend]{algpseudocode}
\algrenewcommand\algorithmicrequire{\textbf{Input:}}
\algrenewcommand\algorithmicensure{\textbf{Output:}}
\def\aligntop#1{\vtop{\null\hbox{#1}}}

\newcommand{\ts}{\textstyle}
\DeclareMathOperator*{\argmax}{arg\,max}

\newcommand{\E}{\mathbb{E}}

\newcommand{\Dir}{\mbox{Dir}}

\newcommand{\hr}{\post{r}}

\newcommand{\htheta}{\post{\theta}}

\newcommand{\cDir}{c_{\text{Dir}}}

\renewcommand{\L}{\mathcal{L}}
\newcommand{\Ldata}{\mathcal{L}_{\text{data}}}
\newcommand{\Lentropy}{\mathcal{L}_{\text{entropy}}}
\newcommand{\Lalloc}{\mathcal{L}_{\text{alloc}}}

\newcommand{\prior}[1]{\bar{#1}}
\newcommand{\post}[1]{\hat{#1}}

\definecolor{keywordcolor}{HTML}{1768EB}
\definecolor{stringcolor}{HTML}{E0142F}
\newcommand{\cprodrho}[1]{\prod_{\ell=1}^{k-1}(1{-}\rho_{\ell})}
\newcommand{\cprodu}[1]{\prod_{\ell=1}^{k-1}(1{-}u_{\ell})}

\title{Fast Learning of Clusters and Topics via Sparse Posteriors}

\author{
Michael C. Hughes and Erik B. Sudderth\\
Department of Computer Science, Brown University, Providence, RI 02912 \\\
\texttt{mike@michaelchughes.com},
\texttt{sudderth@cs.brown.edu}
}

\begin{document} 
\maketitle
\setlength{\abovedisplayskip}{2pt plus 3pt}
\setlength{\belowdisplayskip}{2pt plus 3pt}

\begin{abstract}
Mixture models and topic models generate each observation from a single cluster, but standard variational posteriors for each observation assign positive probability to all possible clusters.
This requires dense storage and runtime costs that scale with the total number of clusters, even though typically only a few clusters have significant posterior mass for any data point.
We propose a constrained family of sparse variational distributions that allow at most $L$ non-zero entries, where the tunable threshold $L$ trades off speed for accuracy.
Previous sparse approximations have used hard assignments ($L=1$), 
but we find that moderate values of $L>1$ provide superior performance.
Our approach easily integrates with stochastic or incremental optimization algorithms to scale to millions of examples. 
Experiments training mixture models of image patches and topic models for news articles show that our approach produces better-quality models in far less time than baseline methods.
\end{abstract}



\section{Introduction}

Mixture models \citep{everitt1981mixtures} and topic models \citep{blei:lda} are fundamental to Bayesian unsupervised learning.
These models find a set of \emph{clusters} or \emph{topics} useful for exploring an input dataset.
Mixture models assume the input data is fully exchangeable, while topic models extend mixtures to handle datasets organized by groups of observations, such as documents or images. 

Mixture and topic models have two kinds of latent variables. \emph{Global} parameters define each cluster, including its frequency and the statistics of associated data.
\emph{Local}, discrete assignments then determine which cluster explains a specific data observation.
For both global and local variables, Bayesian analysts wish to estimate a posterior distribution. 
For these models, full posterior inference via Markov chain Monte Carlo 
(MCMC,~\citet{neal1992bayesianMix}) averages over sampled cluster assignments, producing asymptotically exact estimates at great computational cost.
Optimization algorithms like \emph{expectation maximization} (EM,~\citet{dempster1977maximumLik}) 
or (mean field) \emph{variational Bayes} 
\citep{ghahramani01propagation,winn05variational}
provide faster, deterministic estimates of cluster assignment probabilities. However, at each observation these methods give positive probability to every cluster, requiring dense storage and limiting scalability.


This paper develops new posterior approximations for local assignment variables which allow optimization-based inference to scale to hundreds or thousands of clusters.
We show that adding an additional sparsity constraint to the standard variational optimization objective for local cluster assignments leads to big gains in processing speed. Unlike approaches restricted to hard, winner-take-all assignments, our approach offers a tunable parameter $L$ that determines how many clusters have non-zero mass in the posterior for each observation.
Our approach fits into any variational algorithm,
regardless of whether global parameters are inferred by point estimates (as in EM) or given full approximate posteriors.
Furthermore, our approach 
integrates into existing frameworks for large-scale data analysis
\citep{hoffman:svi,broderick:sva} and is easy to parallelize.
Our open source Python code\footnote{\url{http://bitbucket.org/michaelchughes/bnpy-dev/}}
exploits an efficient C++ implementation of selection algorithms~\citep{blum1973select, musser1997introselect} for scalability.

\begin{figure*}[t!]
\centering
\setlength{\tabcolsep}{0.01cm}
\begin{tabular}{c c c c}
{\scriptsize a: Overall Timings, $K=200$}
&
{\scriptsize b: \textsc{RespFromWeights} step}
&
{\scriptsize c: Summary step}
&
{\scriptsize d: Distance from dense}
\\
\includegraphics[width=0.23\textwidth]{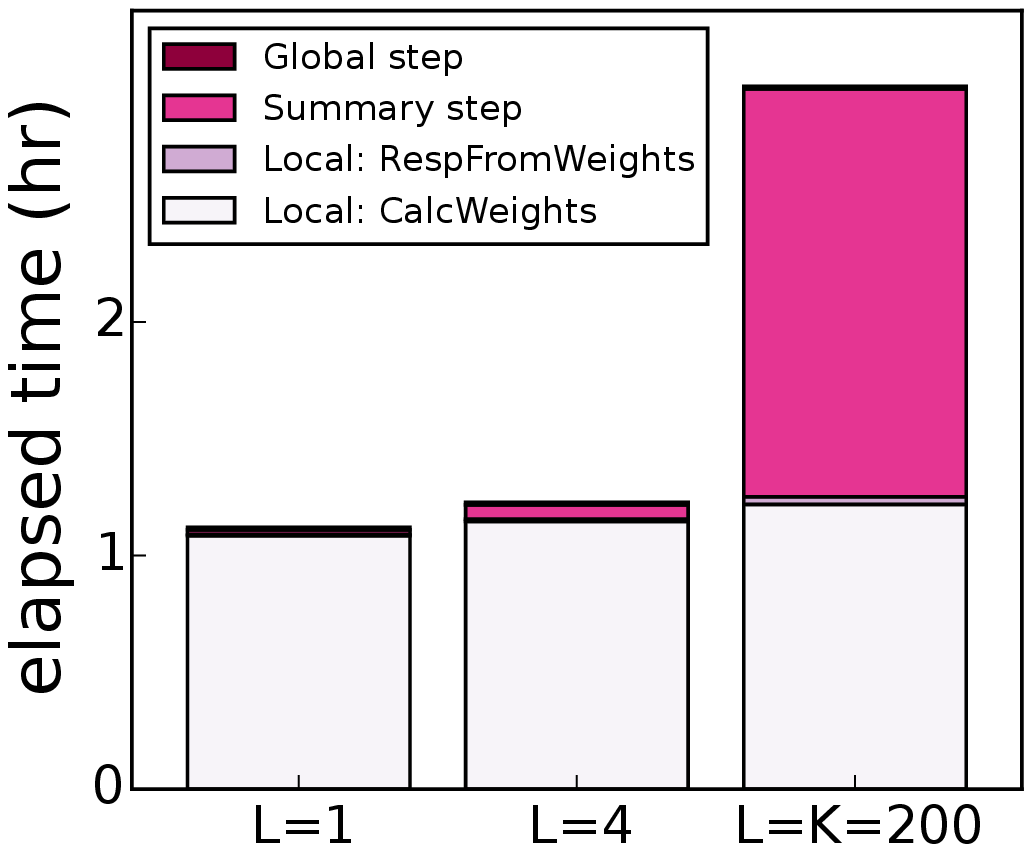}
&
\includegraphics[width=0.23\textwidth]{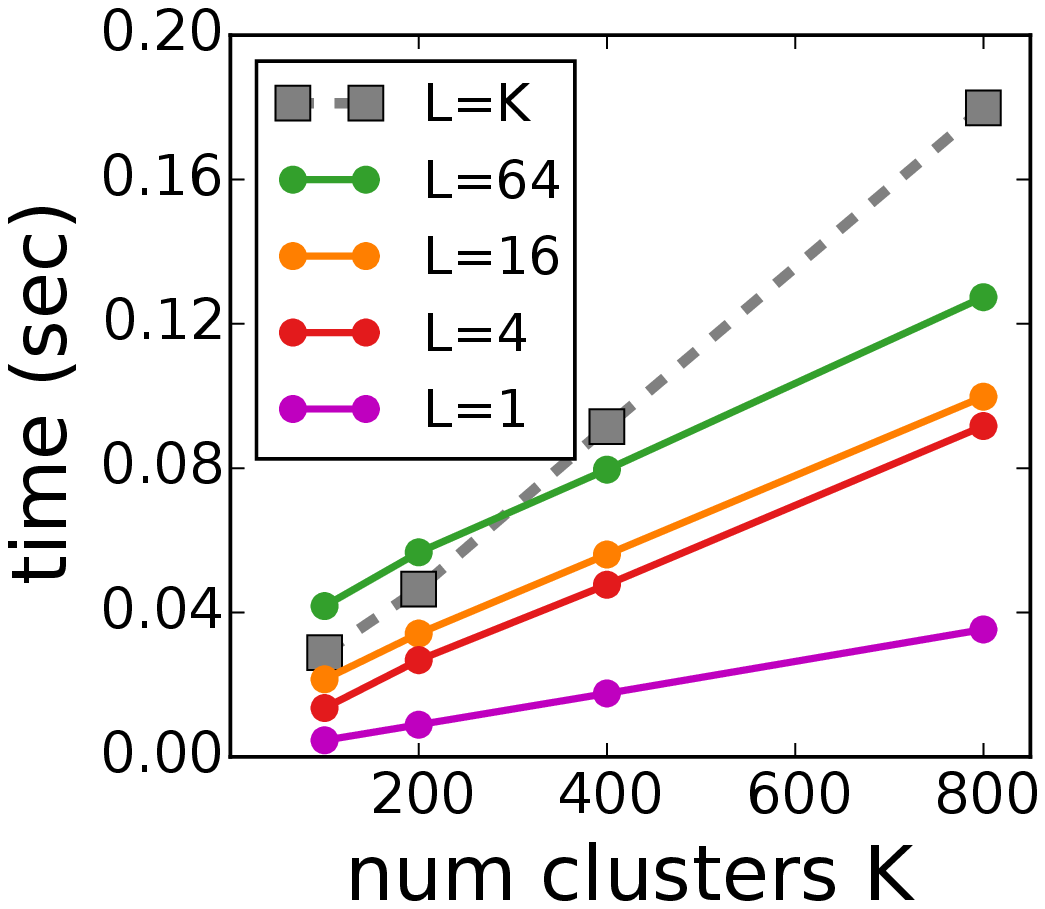}
&
\includegraphics[width=0.23\textwidth]{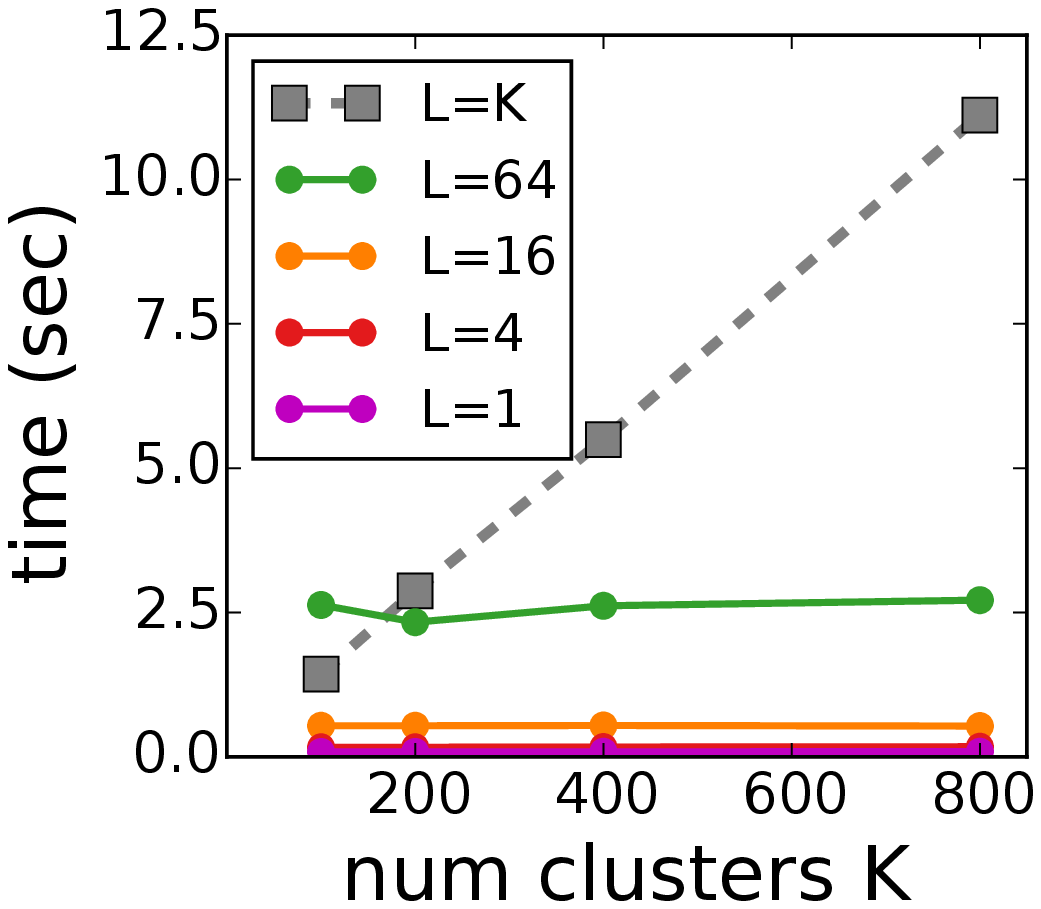}
&
\includegraphics[width=0.23\textwidth]{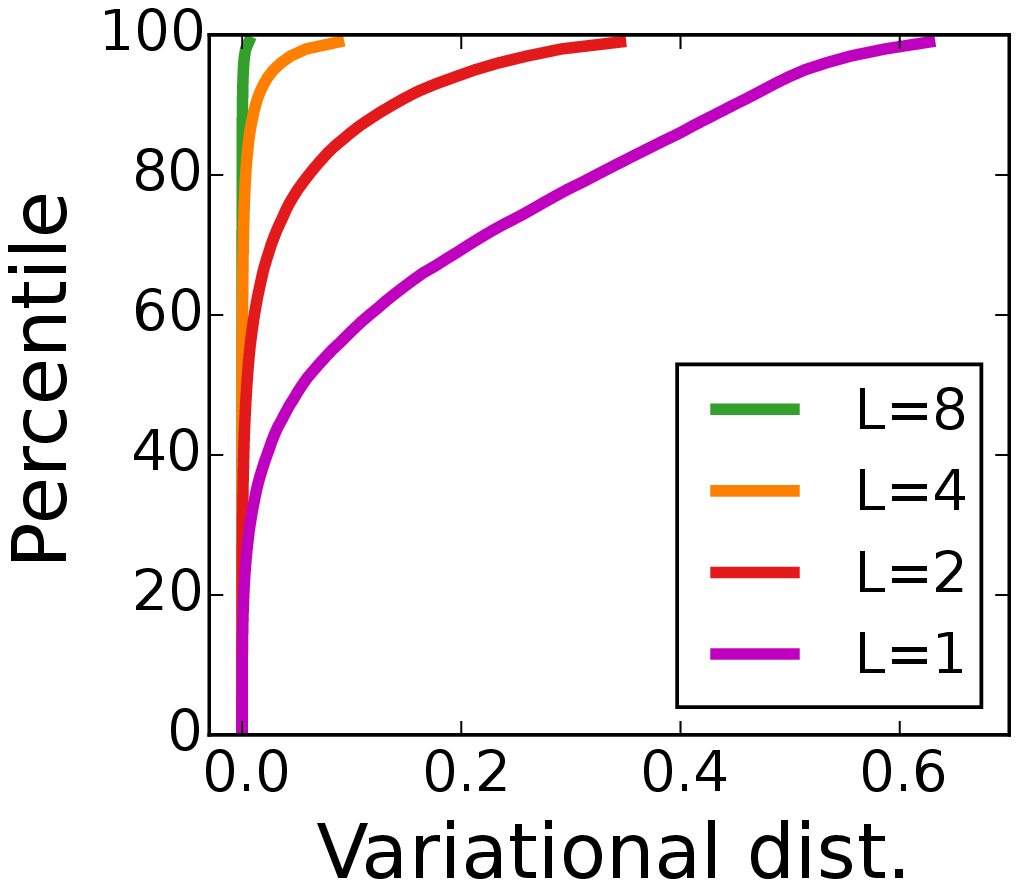}
\end{tabular}
\vspace{-.2cm}
\caption{
Impact of sparsity-level $L$
on the speed and accuracy of estimating a zero-mean Gaussian mixture model for 8x8 pixel patches from natural images.
\emph{a:}
Comparison of algorithm substep costs for 2 complete passes through entire 3.6 million patch dataset from Sec.~\ref{sec:ExperimentsImPatch} with $K=200$ clusters. 
\emph{b:}
Speed comparison of our $L-$sparse approach when computing responsibilities $\hr$ given fixed weights $W$ for 36,000 patches. 
\emph{c:}
Speed comparison of our $L-$sparse approach when computing the per-cluster statistics $\{N_k, S_k \}_{k=1}^K$ defined in Eq.~\eqref{eq:suffStats}
for 36,000 patches.
\emph{d:}
Cumulative density function of variational distance between dense and $L-$sparse responsibilities across 36,000 patches, using the pretrained $K=200$ mixture model published online by \cite{zoran2012natural}.
}
\label{fig:localstepSpeedup}
\end{figure*}

\section{Variational Inference for Mixture Models}
\label{sec:DenseAssignmentMixtures}


Given $N$ observed data vectors $x = \{x_1, x_2, \ldots x_N\}$, a mixture model assumes each observation belongs to one of $K$ clusters. 
Let hidden variable $z_n \in \{1, \ldots, K\}$ denote the specific cluster assigned to $x_n$.  The mixture model has two sets of global parameters, the cluster frequencies $\{\pi_k \}_{k=1}^K$ and cluster shapes $\{ \phi_k \}_{k=1}^K$. 
Let $\pi \sim \mbox{Dir}_K(\alpha/K)$ for scalar $\alpha > 0$,
where $\pi_k$ is the probability of observing data from cluster $k$: 
$z_n \sim \mbox{Cat}_K(\pi)$. We generate observation $x_n$ according to likelihood 
\begin{align}
x_n \sim \mbox{F}( \phi_{z_n} ),
~
\log \mbox{F}(x_n | \phi_k) &= \phi_k^T s(x_n) - c(\phi_k).
\end{align}
The exponential family density $F(x_n | \phi_k)$ has sufficient statistics $s(x_n) \in \mathbb{R}^D$ and natural parameter $\phi_k \in \mathbb{R}^D$. The normalization function $c(\phi_k)$ ensures that $F$ integrates to one. 
We let $\phi_k \sim \mbox{P}(\phi_k | \prior{\lambda})$, where $P$ is a density conjugate to $F$ with parameter $\prior{\lambda}$.
Conjugacy is convenient but not necessary: we only require that the expectation $\E[\log F(x_n | \phi_k)]$ can be evaluated in closed-form.

Mean-field variational inference \citep{wainwright:variational} seeks a factorized posterior $q(z) q(\pi)q(\phi) \approx p(z, \pi, \phi \mid x)$. Each posterior factor has free parameters (denoted with hats) that are optimized to minimize the KL divergence between the simplified approximate density and the true, intractable posterior.
The separate factors for local and global parameters have specially chosen forms:
\begin{align}
q(\pi) &= \mbox{Dir}_K(\pi | \hat{\theta})
, \quad
q(\phi) =
\ts \prod_{k=1}^K \mbox{P}(\phi_k | \hat{\lambda}_k)
, \quad
q(z) = \ts \prod_{n=1}^N \mbox{Cat}(z_n \mid \hat{r}_n).
\end{align}
Our focus is on the free parameter $\hat{r}_n$ which defines the local assignment posterior $q(z_n)$. This vector is non-negative and sums to one. We interpret value $\hat{r}_{nk}$ as the posterior probability of assigning observation $n$ to cluster $k$.
This is sometimes called cluster $k$'s \emph{responsibility} for observation $n$. 

The goal of variational inference is to find the optimal free parameters under a specific objective function $\L$.
Using full approximate posteriors of global parameters 
yields the evidence lower-bound objective function $\L(\hr, \htheta, \hat{\lambda})$ in Eq.~\eqref{eq:MixModelELBOobjective}
which is equivalent to minimizing KL divergence  \citep{wainwright:variational}.
Point estimation of global parameters $\hat{\pi}, \hat{\phi}$ instead yields a maximum-likelihood (ML) objective in Eq.~\eqref{eq:MixModelMAPobjective}.
\begin{align}
\mbox{\small ELBO:~}
\label{eq:MixModelELBOobjective}
\mathcal{L}(x, \hat{r}, \hat{\theta}, \hat{\lambda})
&= \log p(x) - \mbox{KL}(q || p)
=\E_{q(z, \pi, \phi)}[ \log p(x, z, \pi, \phi) - \log q(z, \pi, \phi) ].
\\
\mbox{\small ML:~}
\label{eq:MixModelMAPobjective}
\mathcal{L}(x, \hat{r}, \hat{\pi}, \hat{\phi})
&= \E_{q(z)}[ \log p(x, z \mid \hat{\pi}, \hat{\phi}) - \log q(z)].
\end{align}
Closed-form expressions for both objectives $\L$ are in Appendix~\ref{supp:VariationalForMixtures}. Given an objective function, optimization typically proceeds via coordinate ascent \citep{neal:mem}. We call the update of the data-specific responsiblities $\hr$ the \emph{local} step, which is alternated with the \emph{global} update of $q(\pi),q(\phi)$ or $\hat{\pi},\hat{\phi}$.

\subsection{Computing dense responsiblities during local step}
The local step computes a responsibility vector $\hr_n$ for each observation $n$ that maximizes $\L$ given fixed global parameters.
Under either the approximate posterior treatment of global parameters in Eq.~\eqref{eq:MixModelELBOobjective} or ML objective of Eq.~\eqref{eq:MixModelMAPobjective}, 
the optimal update (dropping terms independent of $\hr_n$) maximizes the following objective function:
\begin{align}
\label{eq:L_r}
\mathcal{L}_n(\hr_n) 
&= \ts \sum_{k=1}^K 
\hr_{nk} W_{nk}(x_n) - \hr_{nk} \log \hr_{nk},
\quad
W_{nk} \triangleq 
\E_q[ \log \pi_k ] + \E_q[ \log F(x_n | \phi_k) ].
\end{align}
We interpret $W_{nk} \in \mathbb{R}$ as the log posterior weight that cluster $k$ has for observation $n$. Larger values imply that cluster $k$ is more likely to be assigned to observation $n$. 
For ML learning, the expectations defining $W_{nk}$ are replaced with point estimates.

Our goal is to find the responsibility vector $\hr_n$ that optimizes $\L_n$ in Eq.~\eqref{eq:L_r}, subject to the constraint that $\hr_n$ is non-negative and sums to one so $q(z_n | \hr_n)$ is a valid density: 
\begin{align}
\label{eq:OptProblemForDenseR}
\hr_{n}^* &= \argmax
\L_n(\hr_n),
\quad \mbox{~subject to~}
\ts \hr_{n} \geq 0, \sum_{k} \hr_{nk} = 1.
\end{align}
The optimal solution is simple: exponentiate each weight and then normalize the resulting vector.
The function \textsc{DenseRespFromWeights} in
Alg.~\ref{alg:RespFromWeights}
details the required steps. 
The runtime cost is $O(K)$, and is dominated by the $K$ required evaluations of the $\mbox{exp}$ function. 

\subsection{Computing sufficient statistics needed for global step}
Given fixed assignments $\hr$, the global step computes the optimal values of the global free parameters under $\mathcal{L}$. 
Whether doing point estimation or approximate posterior inference, this update requires only two finite-dimensional sufficient statistics of $\hr$, rather than all $\hr$ values.
For each cluster $k$, we must compute the expected count $N_k \in \mathbb{R}^+$ of its assigned observations
and the expected data statistic vector $S_k \in \mathbb{R}^D$:
\begin{align}
\ts
N_k(\hr) = \sum_{n=1}^N \hr_{nk} 
, ~~\ts 
S_k(x, \hr) = \sum_{n=1}^N \hr_{nk} s(x_n).
\label{eq:suffStats}
\end{align}
The required work is $O(NK)$ for the count vector and $O(NKD)$ for the data vector.

\newcounter{algcount}
\newcounter{oldfigurecount}
\setcounter{oldfigurecount}{\value{figure}}
\setcounter{figure}{\value{algcount}}
\renewcommand{\figurename}{Alg.}
\begin{figure*}[t!]
\begin{tabular}{c c}
\aligntop{
\begin{minipage}{0.475\textwidth}
  \begin{algorithmic}[1]
\Require{$[W_{n1} \ldots W_{nK}]$ : log posterior weights.}
\Ensure{$[\hr_{n1} \ldots \hr_{nK}]$ : responsibility values
}
\Function{DenseRespFromWeights}{$W_n$}
    \For{$k \in 1, \ldots K$}
        \State $\hr_{nk} = e^{ W_{nk}}$
    \EndFor
    \State $s_n = \ts \sum_{k=1}^K \hr_{nk}$
    \For{$k \in 1, \ldots K$}
        \State $\hr_{nk} = \hr_{nk} / s_n$
    \EndFor
    \State \Return{$\hr_n$}
\EndFunction
  \end{algorithmic}
\end{minipage}
}
&
\aligntop{
\begin{minipage}{0.475\textwidth}
  \begin{algorithmic}[1]
\Require{$[W_{n1} \ldots W_{nK}]$ : log posterior weights.}
\Ensure{
$\{\hr_{n\ell}, i_{n\ell} \}_{\ell=1}^L$ : 
resp. and indices
}
\Function{TopLRespFromWeights}{$W_n$, $L$}
    \State $i_{n1}, \ldots i_{nL} = \textsc{SelectTopL}(W_n)$
    \For{$\ell \in 1, \ldots L$}
        \State $\hr_{n\ell} = e^{ W_{ni_{n\ell}}}$
    \EndFor
    \State $s_n = \ts \sum_{\ell=1}^L \hr_{n\ell}$
    \For{$\ell \in 1, \ldots L$}
        \State $\hr_{n\ell} = \hr_{n\ell} / s_n$
    \EndFor
    \State \Return{$\hr_n, i_n$}
\EndFunction
  \end{algorithmic}
\end{minipage}
}
\end{tabular}
\caption{
Updates for the responsibilities of observation $n$ given weights defined in Eq.~\eqref{eq:L_r}.
\emph{Left:} \textsc{DenseRespFromWeights} is the standard solution to the optimization problem in Eq.~\eqref{eq:OptProblemForDenseR}.
This requires $K$ evaluations of the $\mbox{exp}$ function, $K$ summations, and $K$ divisions.
\emph{Right:} Our proposed method
\textsc{TopLRespFromWeights} optimizes the same objective subject to the additional constraint that at most $L$ clusters have non-zero posterior probability.
First, an $O(K)$ introspective selection algorithm \citep{musser1997introselect} finds the indices of the $L$ largest weights.
Given these, we find the optimum with $L$ evaluations of the $\mbox{exp}$ function, $L$ summations, and $L$ divisions.
}
\label{alg:RespFromWeights}
\end{figure*}

\setcounter{figure}{\value{oldfigurecount}}
\renewcommand{\figurename}{Fig.}

%

\section{Fast Local Step for Mixtures via Sparse Responsibilities}
\label{sec:SparseAssignmentMixtures}
Our key contribution is a new variational objective and algorithm that scales better to large numbers of clusters $K$.
Much of the runtime cost for standard variational inference algorithms comes from representing $\hr_n$ as a dense vector. Although there are $K$ total clusters,
for any observation $n$
only a few entries in $\hr_n$ will have appreciable mass
while the vast majority are close to zero.
We thus further constrain the objective of Eq.~\eqref{eq:OptProblemForDenseR}
to allow at most $1 \leq L \leq K$ non-zero entries:
\begin{align}
 \label{eq:OptProblemForSparseR}
\hr_n^* = \mbox{argmax}_{\hr_n} & \L_n(\hr_n) 
, \quad
\mbox{~s.t.~} \quad
\ts \hr_{n} \geq 0, ~
\sum_{k=1}^K \hr_{nk} = 1,
~ \ts \sum_{k=1}^K \mbox{1}(\hr_{nk} > 0) = L.
\end{align}
The 
function \textsc{TopLRespFromWeights} in Alg.~\ref{alg:RespFromWeights}
solves this constrained optimization problem.
First, we identify the indices of the top $L$ values of the weight vector $W_n$ in descending order. Let $i_{n1}, \ldots, i_{nL}$ denote these top-ranked cluster indices, each one a distinct value in $\{1, 2, \ldots, K\}$.
Given this active set of clusters, we simply exponentiate and normalize only at these indices.
We can represent this solution as an $L$-sparse vector, with $L$ real values $\hr_{n1}, \ldots,\hr_{nL}$ and $L$ integer indices $i_{n1}, \ldots, i_{nL}$.
Solutions are not unique if the posterior weights $W_n$ contain duplicate values. We handle these ties arbitrarily, since swapping duplicate indices leaves the objective unchanged.

\subsection{Proof of optimality.}
\label{sec:proof}
We offer a proof by contradiction that \textsc{TopLRespFromWeights} solves the optimization problem in Eq.~\eqref{eq:OptProblemForSparseR}. Suppose that $\hr_n'$ is optimal, but there exists a pair of clusters $j, k$ such $j$ has larger weight but is not included in the active set while $k$ is. This means $W_{nj} > W_{nk}$, but $\hr'_{nj}=0$ and $\hr'_{nk} > 0$. Consider the alternative $\hr_n^*$ which is equal to vector $\hr_n'$ but with entries $j$ and $k$ swapped. After substituting into Eq.~\eqref{eq:L_r} and simplifying, we find the objective function value increases under our alternative: $\L_n(\hr_n^*) - \L_n(\hr_n') = \hr'_{nk} \cdot (W_{nj} - W_{nk}) > 0$. Thus, the optimal solution must include the largest $L$ clusters by weight in its active set.

\subsection{Runtime cost.}
Alg.~\ref{alg:RespFromWeights} compares \textsc{DenseRespFromWeights} and our new algorithm \textsc{TopLRespFromWeights} side-by-side.
The former requires $K$ exponentiations, $K$ additions, and $K$ divisions to turn weights into responsibilities.
In contrast, given the indices $i_n$ our procedure requires only $L$ of each operation.
Finding the active indices $i_n$ via \textsc{SelectTopL} requires $O(K)$ runtime.

\emph{Selection algorithms} \citep{blum1973select, musser1997introselect} are designed to find the top $L$ values in descending order within an array of size $K$. These methods use divide-and-conquer strategies to recursively partition the input array into two blocks, one with values above a pivot and the other below. \citet{musser1997introselect} introduced a selection procedure which uses \emph{introspection} to smartly choose pivot values and thus guarantee $O(K)$ worst-case runtime.
This procedure is implemented within the C++ standard library as \texttt{nth\_element}, which we use for \textsc{SelectTopL} in practice. This function operates in-place on the provided array, rearranging its values so that the first $L$ entries are all bigger than the remainder. 
Importantly, there is no internal sorting within either partition. Example code is in found in Appendix ~\ref{supp:SelectionAlg}.


Choosing sparsity-level $L$ naturally trades off execution speed and training accuracy. When $L=K$, we recover the original dense responsibilities, while $L=1$ assigns each point to exactly one cluster, as in k-means.
Our focus is on modest values of $1 < L \ll K$.
Fig.~\ref{fig:localstepSpeedup}b shows that for large $K$ values \textsc{TopLRespFromWeights} is faster than \textsc{DenseRespFromWeights} for $L=4$ or $L=16$.
The dense method's required $K$ exponentiations dominates the $O(K)$ introspective selection procedure.

With $L-$sparse responsibilities, computing the statistics $S_k, N_k$ in Eq.~\eqref{eq:suffStats} scales linearly with $L$ rather than $K$.
This gain is useful when applying Gaussian mixture models with unknown covariances to image patches,
where each 8x8 patch requires an expensive 4096-dimensional data statistic $s(x_n) = x_n x_n^T$.
Fig.~\ref{fig:localstepSpeedup}c shows the cost of the summary step virtually disappears when $L=4$ rather than $L=K$. This savings makes the overall algorithm over twice as fast (Fig.~\ref{fig:localstepSpeedup}a), with the remaining bottleneck the dense calculation of weights $W$, which might be sped up for some likelihoods using fast data structures for finding nearest-neighbors.
Fig.~\ref{fig:localstepSpeedup}d shows that $L=8$ captures nearly identical responsibility values as $L=K$, indicating that modest $L$ values may bring speed gains without 
noticeable sacrifice of model quality.

\subsection{Related work.}

\paragraph{Hard assignments.}
One widespread practice used for decades is to consider ``hard'' assignments, 
where each observation is assigned to a single cluster,
instead of a dense vector of $K$ responsibilities.
This is equivalent to setting $L=1$ in our $L-$sparse formulation.
The k-means algorithm \citep{lloyd1982kmeans} 
and its nonparametric extension DP-means \citep{kulis2012dpmeans} justify $L=1$ sparsity via small-variance asymptotics.
So-called ``hard EM'' Viterbi training~\citep{juang1990segmentalKmeans},
or \emph{maximization-expectation} algorithms~
\citep{kurihara2009ME} both use $L=1$ hard assignments.
However, we expect $L=1$ to be too coarse for many applications while moderate values like $L=8$ offer better approximations, as shown in Fig.~\ref{fig:localstepSpeedup}d.

\paragraph{Sparse EM.}
A prominent early method to exploit sparsity in responsibilities is the Sparse EM algorithm proposed by \cite{neal:mem}. Sparse EM maintains a dense vector $\hr_n$ for each observation $n$, but only edits a subset of this vector during each local step. The edited subset may consist of the $L$ largest entries or all entries above some threshold. Any inactive entries are ``frozen'' to current non-zero values and newly edited entries are normalized such that the length-$K$ vector $\hr_n$ preserves its sum-to-one constraint. 

Sparse EM can be effective for small datasets with a few thousand examples and has found applications such as MRI medical imaging~\citep{Ng2004sparseEMforMRI}. 
However, our $L-$sparse approach has three primary advantages relative to Sparse EM:
(1) Our $L-$sparse method requires less per-observation memory for responsibilities. While Sparse EM must store $K$ floating-point values to represent a responsibility vector, we need to store only $L$.
(2) Our $L-$sparse method easily scales to minibatch-based training algorithms in Sec.~\ref{sec:scalable_algorithms}, but Sparse EM's required storage is prohibitive. Our approach can safely discard responsiblity vectors after required sufficient statistics are computed. Sparse EM must explicitly store responsibilities for every observation in the dataset at cost $O(NK)$ if future sparse updates are desired. This prohibits scaling to millions of examples by processing small minibatches, unless each minibatch has its full responsibility array written to and from disk when needed.
(3) We proved in Sec.~\ref{sec:proof} that top-L selection is the optimal way to compute $L-$sparse responsibilities and monotonically improve our training objective function. \cite{neal:mem} suggest this selection method only as a heuristic without justification.

\paragraph{Expectation Truncation.}
When we undertook most of this research, we were unaware of a related method by \cite{lucke2010expectation} called \emph{Expectation Truncation} which constrains the approximate posterior probabilities of discrete or multivariate binary variables to be $L-$sparse. 
\cite{lucke2010expectation} considered non-negative matrix factorization and sparse coding problems.
Later extensions applied this core algorithm to mixture-like sprite models for cleaning images of text documents  \citep{dai2014autonomous} and spike-and-slab sparse coding \citep{sheikh2014truncated}. Our work is the first to apply $L-$sparse ideas to mixture models and topic models.

The original Expectation Truncation algorithm \citep[Alg. 1]{lucke2010expectation} expects a user-defined selection function to identify the entries with non-zero responsibility for a specific observation. In practice, the selection functions they suggest are chosen heuristically, such as the upper bound in Eq. 28 of \citep{lucke2010expectation}. The original authors freely admit these selection functions are not optimal and may not monotonically improve the objective function \citep[p. 2869]{lucke2010expectation}. In contrast, we proved in Sec.~\ref{sec:proof} that top-$L$ selection will optimally improve our objective function.

One other advantage of our work over previous Expectation Truncation efforts are our thorough experiments exploring how different $L$ values impact training speed and predictive power. Comparisons over a range of possible $L$ values on real datasets are lacking in \cite{lucke2010expectation} and other papers. Our key empirical insight is that modest values like $L=4$ are frequently better than $L=1$, especially for topic models. 

%

\subsection{Scalabilty via minibatches}
\label{sec:scalable_algorithms}

\paragraph{Stochastic variational inference (SVI).}
Introduced by \cite{hoffman:olda}, SVI scales standard coordinate ascent to large datasets by processing subsets of data at a time. Our proposed sparse local step fits easily into SVI.
At each iteration $t$, SVI performs the following steps:
(1) sample a batch $\mathcal{D}^t \subset \{x_1, \ldots x_N\}$ from the full dataset, uniformly at random; (2) for each observation $n$ in the batch, do a local step to update responsibilities $\hr_{n}$ given fixed global parameters; (3) update the global parameters by stepping from their current values in the direction of the natural gradient
of the rescaled batch objective $\L(\mathcal{D}^t)$. This procedure is guaranteed to reach a local optima of $\L$ if the step size of the gradient update decays appropriately as $t$ increases \citep{hoffman:svi}.

\paragraph{Incremental algorithms (MVI).}
Inspired by incremental EM \citep{neal:mem}, \cite{hughes:moVB} introduced \emph{memoized} variational inference (MVI). The data is divided into a fixed set of $B$ batches before iterations begin.
Each iteration $t$ completes four steps: (1) select a single batch $b$ to visit; (2) for each observation $n$ in this batch, compute optimal local responsibilities $\hr_n$ given fixed global parameters and summarize these into sufficient statistics for batch $b$; (3) incrementally update 
s
whole-dataset statistics given the new statistics for batch $b$; (4) 
compute optimal global parameters given the whole-dataset statistics.
The incremental update in step (3) requires caching (or ``memoizing'') the summary statistics in Eq.~\eqref{eq:suffStats} at each batch.
This algorithm has the same per-iteration runtime as stochastic inference, but guarantees the monotonic increase of the objective $\L$ when the local step has a closed-form solution like the mixture model.
Its first pass through the entire dataset is equivalent to streaming variational Bayes~\citep{broderick:sva}.


\section{Mixture Model Experiments}
\label{sec:ExperimentsImPatch}
We evaluate dense and $L-$sparse mixture models for natural images, inspired by \cite{zoran2012natural}. We train a model for 8x8 image patches taken from overlapping regular grids of stride 4 pixels.
Each observation is a vector $x_n \in \mathbb{R}^{64}$, preprocessed to remove its mean.
We then apply a mixture model with zero-mean, full-covariance Gaussian likelihood function $F$. We set concentration $\alpha = 10$.
To evaluate, we track the log-likelihood score of heldout observations $x'_n$ under our trained model, defined as
$\log p(x'_n) = \ts
\log \sum_{k=1}^K \hat{\pi}_k \mathcal{N}( x'_n | 0, \hat{\Sigma}_k )
$.
Here, $\hat{\pi}_k = \E_q[ \pi_k]$ and $\hat{\Sigma}_k = \E_q[\Sigma_k]$ are point estimates computed from our trained global parameters using standard formulas. 
The function $\mathcal{N}$ is the probability density function of a multivariate normal.

\begin{figure*}[!t]
\includegraphics[width=\textwidth]{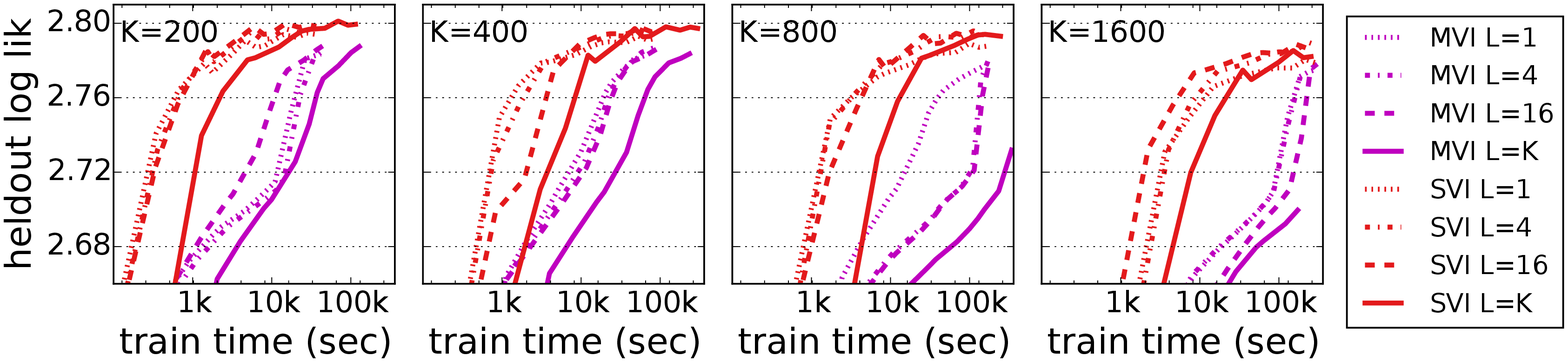}
\vspace{-.3cm}
\caption{
Analysis of 3.6 million 8x8 pixel image patches using a zero-mean Gaussian mixture model trained by $L-$sparse stochastic (SVI) and memoized (MVI) variational algorithms.
We train on 400 total images processed 4 images at a time.
Each panel shows the heldout log likelihood score over time for training runs with various sparsity levels at a fixed number of clusters $K$. Training time is plotted on log-scale.
}
\label{fig:results-ImgPatch}
\end{figure*}

Fig.~\ref{fig:results-ImgPatch}
compares $L-$sparse implementations of SVI and MVI on 3.6 million patches from 400 images. The algorithms process $100$ minibatches each with $N=36,\!816$ patches.
We see the sparse methods consistently reach good predictive scores 2-4 times faster than dense $L=K$ runs do (note the log-scale of the time axis).
Finally, modestly sparse $L=16$ runs often reach higher values of heldout likelihood than hard $L=1$ runs, especially in the $K=800$ and $K=1600$ plots for SVI (red).

%
\begin{figure*}[t!]
\centering
\setlength{\tabcolsep}{.01cm}
\begin{tabular}{c c c c}
\scriptsize{a: Overall Timings, K=800}
&
\scriptsize{b: Local Step + Restarts}
&
\scriptsize{c: Summary Step}
&
\scriptsize{d: Distance from $L=K$}
\\
\includegraphics[width=0.27\textwidth]{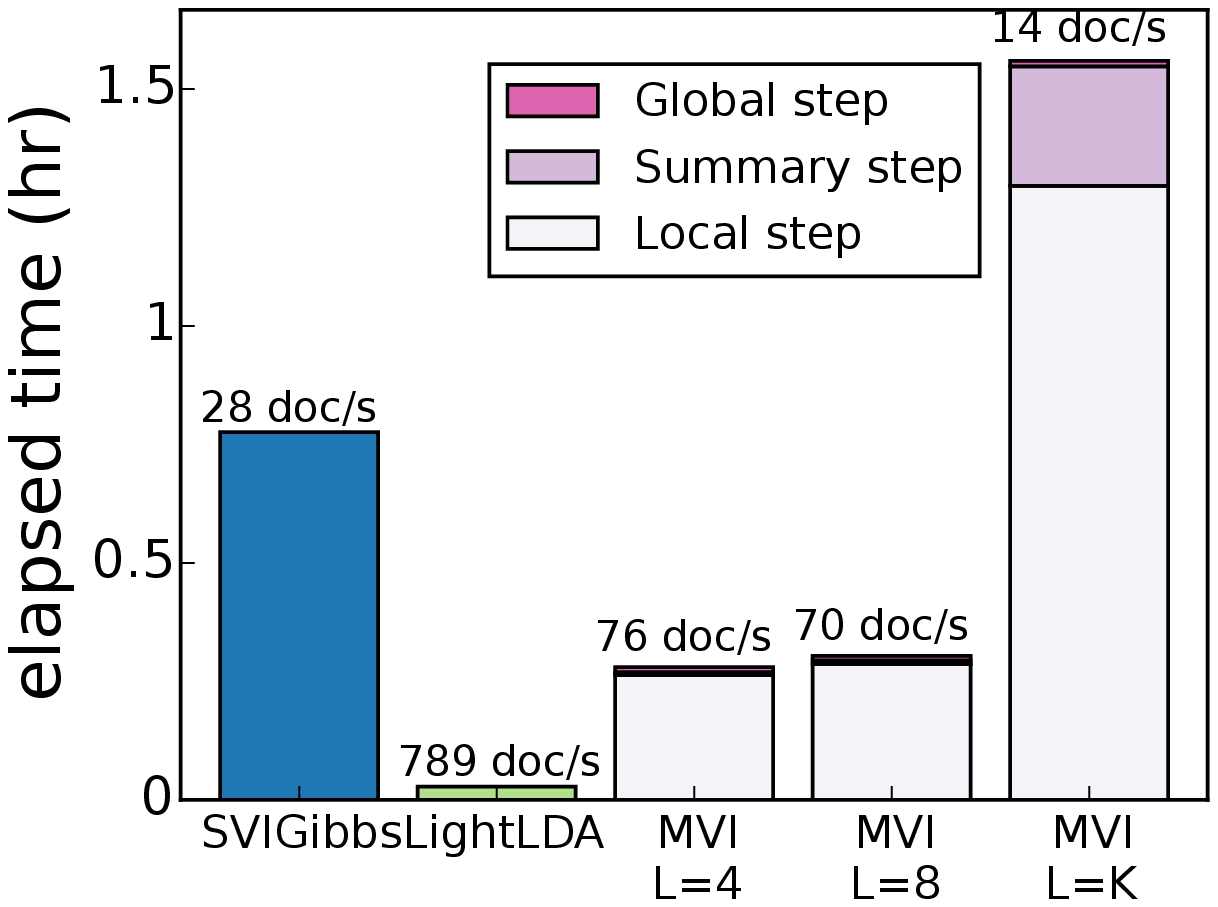}
&
\includegraphics[width=0.22\textwidth]
{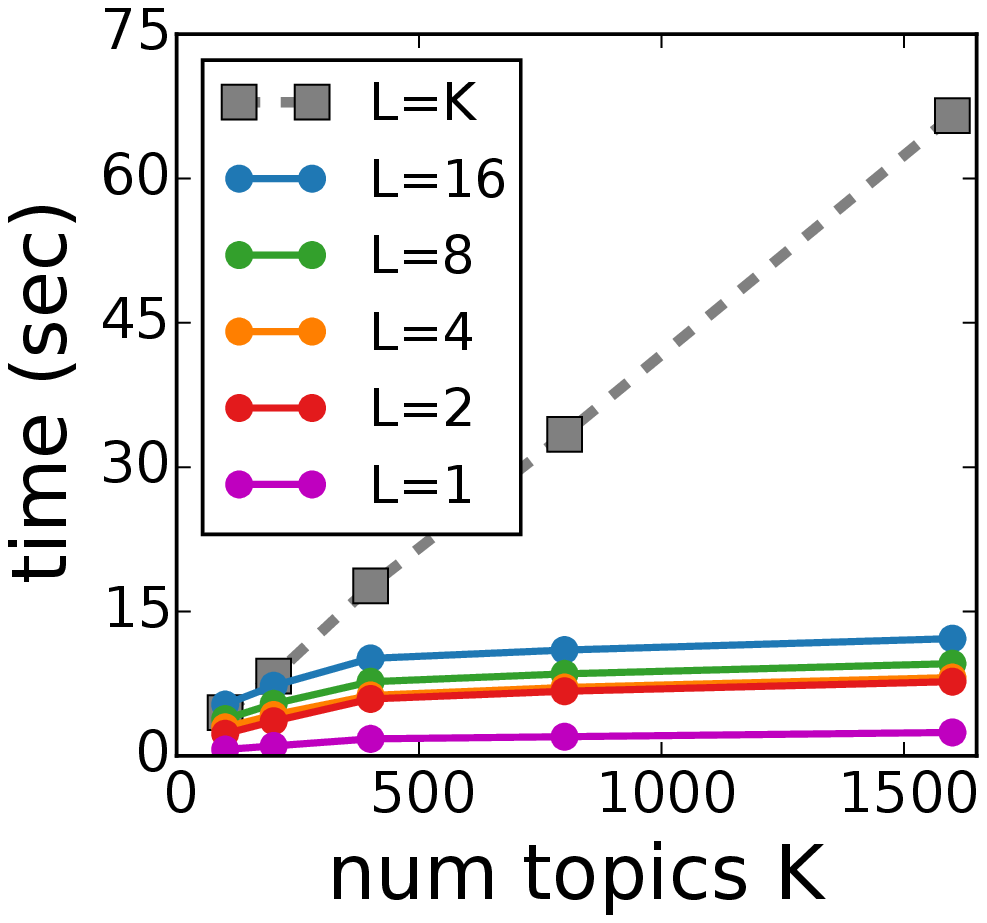}
&
\includegraphics[width=0.22\textwidth]{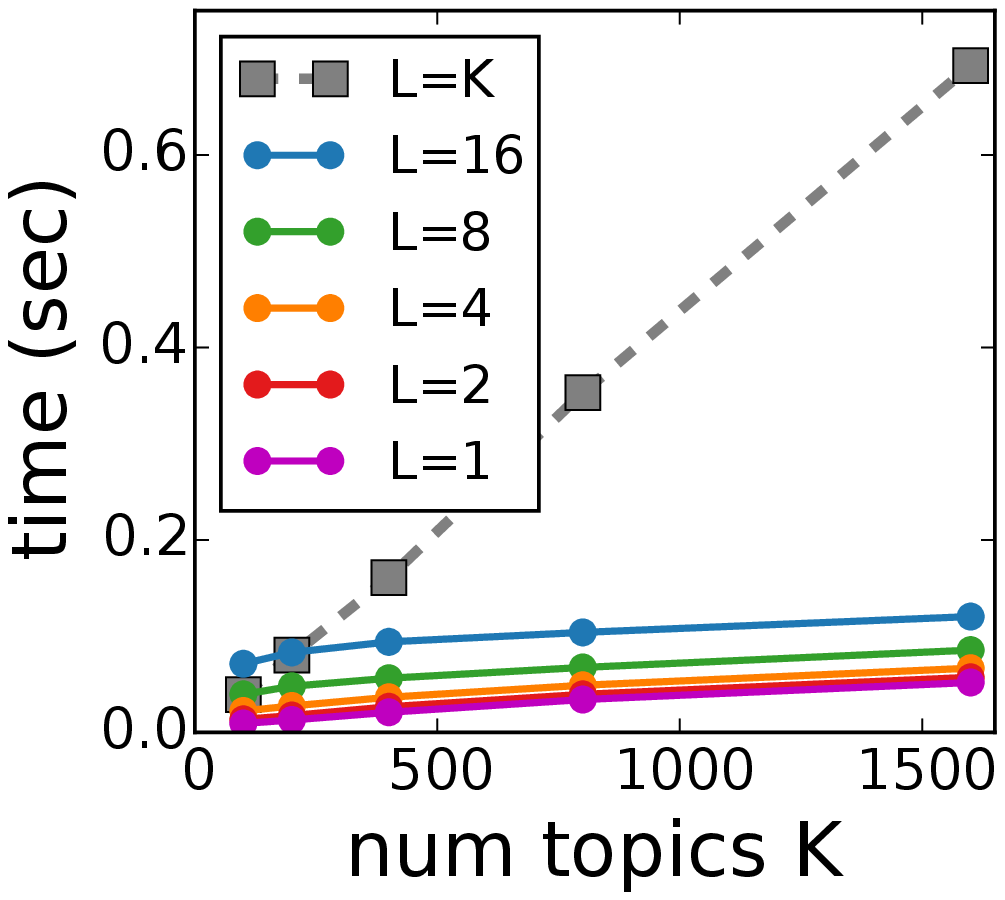}
&
\includegraphics[width=0.22\textwidth]{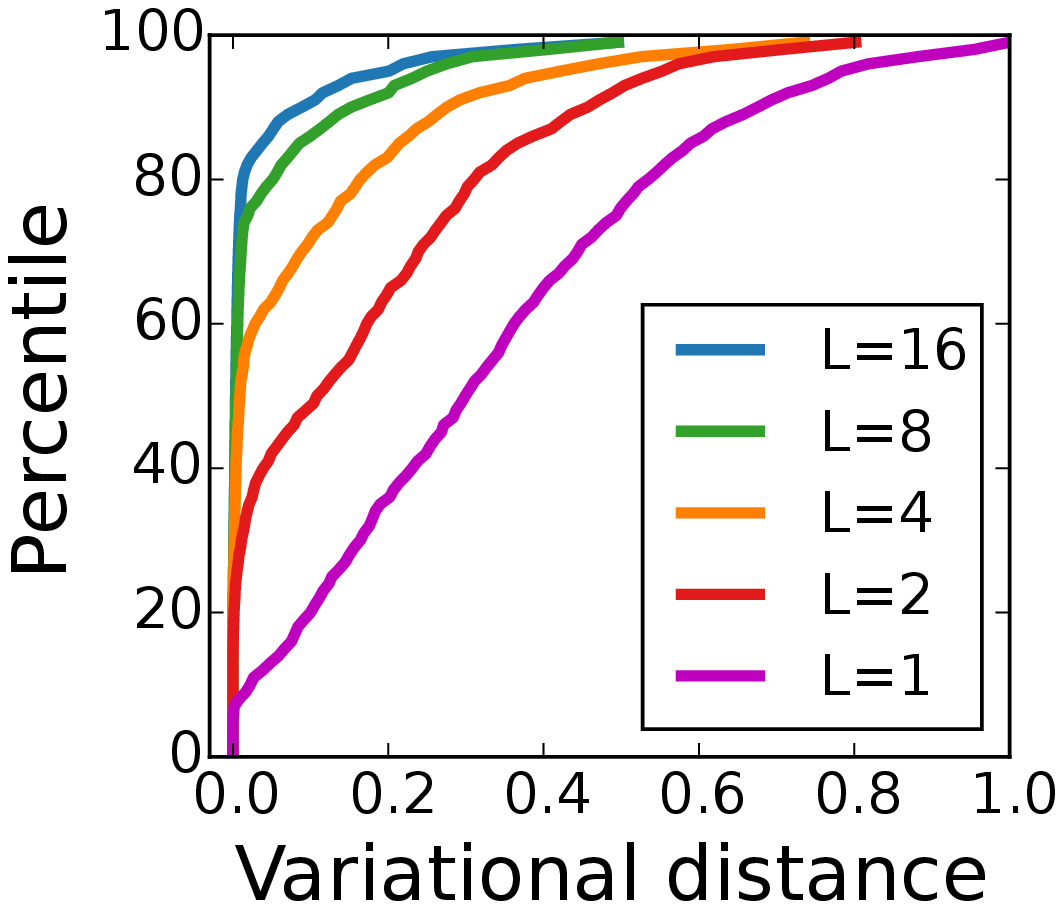}
\end{tabular}
\vspace{-.2cm}
\caption{
Impact of sparsity-level $L$ on speed and accuracy of training topic models.
\emph{a:}
Comparison of the runtime costs for 10 complete passes through 7981 Wikipedia documents with $K=800$ topics.
Our $L-$sparse methods further show breakdown by algorithm substeps.
Text above each bar indicates the number of documents processed per second.
\emph{b:}
Timings for the $L-$sparse local step with restart proposals on 1000 NYTimes articles, using 100 iterations at each document.
\emph{c:}
Timings for the $L$-sparse summary step on 1000 NYTimes articles.
\emph{d:}
Cumulative density function of variational distance between dense and $L-$sparse document-topic distributions, across 1000 NYTimes documents.
We define the empirical topic distribution of document $d$ by normalizing the count vector $[N_{d1} \ldots N_{dK}]$. 
}
\label{fig:TopicModelLocalStepSpeedup}
\end{figure*}

\section{Fast Local Step for Topic Models via Sparse Responsibilities}
\label{sec:TopicModels}

We now develop a sparse local step for topic models.
Topic models \citep{blei2012topicmodels} are hierarchical mixtures applied to discrete data from $D$ documents, $x_1, \ldots x_D$. 
Let each document $x_d$ consist of observed word tokens 
from a fixed vocabulary of $V$ word types, 
though we could easily build a topic model for observations of any type (real, discrete, etc.).
Each document $d$ contains $N_d$ observed word tokens
$x_d = \{ x_{dn} \}_{n=1}^{N_d}$,
where token $x_{dn} \in \{1, \ldots V\}$ identifies the type of the $n$-th word. 

The \emph{latent Dirichlet allocation} (LDA) topic model \citep{blei:lda}
generates a document's observations from a mixture model with common topics $\{\phi\}_{k=1}^K$ but document-specific frequencies $\pi_d$.
Each topic $\phi_{k} \sim \mbox{Dir}_V(\prior{\lambda})$, where $\phi_{kv}$ is the probability of type $v$ under topic $k$.
The document-specific frequencies $\pi_d$ are drawn from a symmetric Dirichlet $\mbox{Dir}_K(\frac{\alpha}{K} \ldots \frac{\alpha}{K})$, where $\alpha > 0$ is a scalar. Assignments are drawn $z_{dn} \sim \mbox{Cat}_K(\pi_d)$, 
and then the observed words are drawn
$x_{dn} \sim \mbox{Cat}_V(\phi_{z_{dn}})$.

The goal of posterior inference is to estimate the common topics as well as the frequencies and assignments in any document.
The standard mean-field approximate posterior \citep{blei:lda} is:
\begin{align}
q(z_{d}) &= \ts \prod_{n=1}^{N_d}
\mbox{Cat}_K(z_{dn} | \hr_{dn})
, \quad
q(\pi_d) = \mbox{Dir}_K(\pi_d | \htheta_{d})
, \quad
q(\phi) = \ts \prod_{k=1}^K \mbox{Dir}_V(\phi_k | \post{\lambda}_{k}).
\end{align}
Under this factorization, we again set up a standard optimization objective $\L(x, \hr, \htheta, \hat{\lambda})$ as in Eq.~\eqref{eq:MixModelELBOobjective}. Complete expressions are in Appendix~\ref{supp:VariationalForLDA}.
We optimize this objective via coordinate ascent, 
alternating between local and global steps.
Our focus is the local step, which requires updating both the assignment factor $q(z_d | \hr_d)$ and the frequencies factor $q(\pi_d | \htheta_d)$ for each document $d$.
Next, we derive an interative update algorithm for estimating the assignment factor $q(z_d)$ and the frequencies factor $q(\pi_d)$ for a document $d$. Alg.~\ref{alg:TopicModelLocalStep} lists the conventional algorithm and our new sparse version.

\paragraph{Document-topic update.}
Following \citep{blei:lda}, we have a closed-form update for each topic $k$: $\htheta_{dk} \gets N_{dk}(\hr_d) + \alpha/K$.
This assumes that responsibilities $\hr_{d}$ have been summarized into counts $N_{dk}(\hr_d)$ of the 
number of tokens assigned to topic $k$ in document $d$:
$N_{dk} \triangleq \sum_{n=1}^N \hr_{dnk}$.

\paragraph{Responsibility update.}
As in \citep{blei:lda}, the optimal update for the dense responsibilities $\hr_{dn}$ for token $n$ has a closed form like the mixture model, but with document-specific weights:
\begin{align}
\label{eq:Update_r_dn_TopicModel}
\hr_{dn} \gets \textsc{DenseRespFromWeights}(W_{dn})
, \quad 
W_{dnk} &\triangleq \E_q[ \log \pi_{dk} + \log \phi_{kx_{dn}} ],
\\ \notag 
\E_q[\log \pi_{dk}] &\triangleq 
\ts \psi(\htheta_{dk}) 
\ts - \psi(\sum_{\ell=1}^K \htheta_{d\ell}).
\end{align}
We can incorporate our $L$-sparse constraint from Eq.~\eqref{eq:OptProblemForSparseR}
to obtain sparse rather than dense responsibilties.
The procedure \textsc{TopLRespFromWeights} from Alg.~\ref{alg:RespFromWeights} still provides the optimal solution.
%

\paragraph{Iterative joint update for dense case.}
Following standard practice for dense assignments \citep{blei:lda}, we use a block-coordinate ascent algorithm that iteratively updates $\hr_d$ and $\htheta_d$ using the closed-form steps above. 
To initialize the update cycle, we recommend setting the initial weights as if the document-topic frequencies are uniform: $W_{dnk} = \E_q[ \log \phi_{k x_{dn}}] + \log \frac{1}{K}$. This lets the topic-word likelihoods drive the initial assignments. We then alternate updates until either a maximum number of iterations is reached (typically 100) or the maximum change in document-topic counts $N_{dk}$ falls below a threshold (typically 0.05).
Appendix~\ref{supp:TopicModelLocalStep} provides a detailed algorithm.
Fig.~\ref{fig:TopicModelLocalStepSpeedup}a
compares the runtime cost of the local, summary, and global steps of the topic model, showing that the local iterations dominate the overall cost. 

\paragraph{Iterative joint update with sparsity.}
Our new $L$-sparse constraint on responsibilities leads to a fast local step algorithm for topic models.
This procedure has two primary advantages over the dense baseline.
First, we use \textsc{TopLRespFromWeights} to update the per-token responsibilities $\hr_{dn}$, resulting in faster updates.
Second, we further assume that once a topic's mass $N_{dk}$ decays near zero, it will never rise again. With this assumption,
at every iteration we identify the set of active topics (those with non-neglible mass) in the document: $\mathcal{A}_d \triangleq \{k : N_{dk} > \epsilon \}$.
Only these topics will have weight large enough to be chosen in the top $L$ for any token.
Thus, throughout local iterations we consider only the active set of topics, reducing all steps from cost $O(K)$ to cost $O(|\mathcal{A}_d|)$.

Discarding topics within a document when mass becomes very small is justified by previous empirical observations of the ``digamma problem'' described in \cite{mimno2012sparsesvi}: for topics with negligible mass, the expected log prior weight $E[\log \pi_{dk}]$ becomes vanishingly small. For example, $\psi(\frac{\alpha}{K}) \approx -200$ for $\alpha \approx 0.5$ and $K \approx 100$, and gets smaller as $K$ increases.
In practice, after the first few iterations the active set stabilizes and each token's top $L$ topics rarely change while the relative responsibilities continue to improve. In this regime, we can reduce runtime cost by avoiding selection altogether, instead just reweighting each token's current set of top $L$ topics.
We perform selection for the first 5 iterations and then only every 10 iterations, which yields large speedups without loss in quality.

Fig.~\ref{fig:TopicModelLocalStepSpeedup} 
compares the runtime of our sparse local step across values of sparsity-level $L$
against a 
comparable implementation of the standard dense algorithm. 
Fig.~\ref{fig:TopicModelLocalStepSpeedup}b shows that our $L-$sparse local step can be at least 3 times faster when $K=400$. Larger $K$ values lead to even larger gains.
Fig.~\ref{fig:TopicModelLocalStepSpeedup}c shows that sparsity improves the speed of the summary step, though this step is less costly than the local step for topic models.
Finally, Fig.~\ref{fig:TopicModelLocalStepSpeedup}d shows that modest $L=8$ sparsity yields document-topic distributions very close to those found by the dense local step, while $L=1$ is much coarser.

\paragraph{Restart proposals.}
In scalable applications, we assume that we cannot afford to store any document-specific information between iterations. 
Thus, each time we visit a document $d$ we must infer both $q(\pi_d | \htheta_d)$ and $q(z_d | \hr_d)$ from scratch. This joint update is non-convex, and thus our recommended cold-start initialization for $\htheta_d$ is \emph{not} guaranteed to monotonically improve $\L$ across repeat visits to a document. However, even if we could store document counts across iterations we find warm-starting often gets stuck in poor local optima (see Fig.~\ref{fig:warm_vs_cold} of Appendix~\ref{supp:TopicModelLocalStep}).
Instead, we combine cold-starting with \emph{restart} proposals.
\cite{hughes:hdpreliable} introduced
restarts as a post-processing step 
for the single document local iterations that 
results in solutions $\hr_d, \htheta_d$ with better objective function scores.
Given some fixed point $(\hr_d, \htheta_d)$, the restart proposal constructs a candidate $(\hr_d', \htheta_d')$ by 
forcing all responsibility mass on some active topic to zero
and then running a few iterations forward.
We accept the new proposal if it improves the objective $\L$. These proposals escape local optima by finding nearby solutions which favor the prior's bias toward sparse document-topic probabilities. They are frequently accepted in practice (40-80\% in a typical Wikipedia run), so we always include them in our sparse and dense local steps.

\paragraph{Related work.}
MCMC methods specialized to topic models of text data can exploit sparsity for huge speed gains.
SparseLDA~\citep{yao2009SparseLDA} is a clever decomposition of the Gibbs conditional distribution to make each per-token assignment step cost less than $O(K)$.
AliasLDA~\citep{li2014aliaslda} and LightLDA~\citep{yuan2015lightlda} both further improve this to amortized $O(1)$.
These methods are still limited to hard assignments and are only applicable to discrete data. In contrast, our approach allows expressive intermediate sparsity and can apply to a broader family of mixtures and topic models for real-valued data.

More recently, several efforts have used MCMC samplers 
to approximate the local step within a larger variational algorithm
\citep{mimno2012sparsesvi,wang:trunc_free}. 
They estimate an approximate posterior $q(z_d)$ by averaging over many samples, where each sample is an $L=1$ hard assignment.
The number of finite samples $S$ needs to be chosen to balance accuracy and speed.
In contrast, our sparsity-level $L$ provides more intuitive control over approximation accuracy and optimizes $\L$ exactly, not just in expectation.

\setcounter{oldfigurecount}{\value{figure}}
\stepcounter{algcount}
\setcounter{figure}{\value{algcount}}
\renewcommand{\figurename}{Alg.}
\begin{figure*}[t!]
\setlength{\tabcolsep}{0.01cm}
\begin{tabular}{c c}
\aligntop{
\begin{minipage}{0.49\textwidth}
  \begin{algorithmic}[1]
\Require{
    \Statex $\alpha$ : document-topic smoothing scalar
    \Statex
        $\{\{ C_{vk} \}_{k=1}^K\}_{v=1}^V$ : log prob. of word $v$ in topic $k$
    \Statex
        \hspace{2cm} $C_{vk} \triangleq \E_q[ \log \phi_{kv}]$
    \Statex
        $\{v_{du},c_{du}\}_{u=1}^U$ : word type/count pairs for doc.~$d$
    \Statex ~
}
\Ensure{
    \Statex
        $\hr_{d}$ : dense responsibilities for doc~$d$
    \Statex
        $[\htheta_{d1} \ldots \theta_{dK}]$ : topic pseudo-counts for doc~$d$
}
\Function{DenseStepForDoc}{$C, \alpha, v_d, c_d$}
\For{$u = 1, \ldots U$}
    \State $\hr_{du} = $ {\footnotesize\textsc{DenseRespFromWeights}}$(C_{v_{du}})$
\EndFor
\While{not converged}
    \For{$k = 1,2 \ldots K$}
        \State $N_{dk} = \ts \sum_{u} c_{du} \hr_{duk}$
        \State $P_{dk} = \psi(N_{dk} + \frac{\alpha}{K})$ \Comment{Implicit $\htheta_d$}
    \EndFor
    \For{$u = 1, 2 \ldots U$}
        \For{$k = 1, 2 \ldots K$}
            \State $W_{duk} = C_{v_{du}k} + P_{dk}$
        \EndFor
        \State $\hr_{du} = ${\footnotesize\textsc{DenseRespFromWeights}}$(W_{du})$   
    \EndFor
\EndWhile
\For{$k = 1,2 \ldots K$}
    \State $N_{dk} = \ts \sum_{u} c_{du} \hr_{duk}$
    \State $\htheta_{dk} = N_{dk} + \frac{\alpha}{K}$
\EndFor
\State \Return $\hr_d, \htheta_d$
\EndFunction
  \end{algorithmic}
\end{minipage}
}
&
\aligntop{
\begin{minipage}{0.49\textwidth}
  \begin{algorithmic}[1]
\Require{
    \Statex $\alpha$ : document-topic smoothing scalar
    \Statex
        $\{\{ C_{vk} \}_{k=1}^K\}_{v=1}^V$ : log prob. of word $v$ in topic $k$
    \Statex
        \hspace{2cm} $C_{vk} \triangleq \E_q[ \log \phi_{kv}]$
    \Statex
        $\{v_{du},c_{du}\}_{u=1}^U$ : word type/count pairs for doc.~$d$
    \Statex $L$ : integer sparsity level
}
\Ensure{
    \Statex $\hr_{d}, i_d$ : $L$-sparse responsibilities and indices
    \Statex
        $[\htheta_{d1} \ldots \theta_{dK}]$ : topic pseudo-counts for doc~$d$
}
\Function{LSparseStepForDoc}{$C, \alpha, v_d, c_d, L$}
\For{$u = 1, \ldots U$}
    \State $\hr_{du}, i_{du} {=} ${\footnotesize \textsc{TopLRespFromW}}$(C_{v_{du}}, L)$
\EndFor
\For{$k = 1, \ldots K$}
    \State $N_{dk} = \ts \sum_{u=1}^U c_{du} \hr_{duk}$
\EndFor
\State $\mathcal{A}_d = \{ k \in [1, K] : N_{dk} > \epsilon \}$
\While{not converged}
    \For{$k \in \mathcal{A}_d$}
        \State $P_{dk} = \psi(N_{dk} + \frac{\alpha }{K})$
    \EndFor
    \For{$u = 1, 2 \ldots U$}
        \For{$k \in \mathcal{A}_d$}
            \State $W_{duk} = C_{v_{du}k} + P_{dk}$
        \EndFor
        \State $\hr_{du}, i_{du} = ${\footnotesize \textsc{TopLRespFromW}}$(W_{du}, L)$
    \EndFor
    \For{$k \in \mathcal{A}_d$}
        \State $N_{dk} = \ts \sum_{u=1}^U c_{du} \hr_{duk}$
    \EndFor
    \State $\mathcal{A}_d = \{ k \in \mathcal{A}_d : N_{dk} > \epsilon \}$
\EndWhile
\For{$k = 1,2 \ldots K$}
    \State $\htheta_{dk} = N_{dk} + \frac{\alpha}{K}$
\EndFor
\State \Return $\hr_d, i_d, \htheta_d$
\EndFunction
\end{algorithmic}
\end{minipage}
}
\end{tabular}
\caption{
Algorithms for computing the per-unique-token responsibilities $\{ \hr_{du} \}_{u=1}^{U_d}$ and topic pseudocounts $\htheta_d$ for a single document $d$ given a fixed set of $K$ topics.
\emph{Left:} In the standard dense algorithm, each step scales linearly with the number of total topics $K$, regardless of how many topics are used in the document.
\emph{Right:} 
In our $L-$sparse algorithm, forcing each observation to use at most $L$ topics and tracking the active topics in a document $\mathcal{A}_d$ leads to update steps that scale linearly with the number of active topics $|\mathcal{A}_d|$, which can be much less than the total number of topics $K$.
}
\label{alg:TopicModelLocalStep}
\end{figure*}
\setcounter{figure}{\value{oldfigurecount}}
\renewcommand{\figurename}{Fig.}

%

\section{Topic Model Experiments}
\label{sec:ExperimentsTopicModels}
We compare our $L-$sparse implementations of MVI and SVI to external baselines: 
SparseLDA~\citep{yao2009SparseLDA}, a fast implementation of standard
Gibbs sampling~\citep{griffiths:2004:fst}; and SVIGibbs~\citep{mimno2012sparsesvi}, a stochastic variational method that uses Gibbs sampling to approximate local gradients.
These algorithms use Java code from Mallet \citep{MALLET}.
We also compare to the public C++ implementation of LightLDA \citep{yuan2015lightlda}.
External methods use their default initialization, while we sample $K$ diverse documents using the Bregman divergence extension~\citep{ackermann2009bregmanCoresets} of k-means++~\citep{arthur:kmeansplusplus} to initialize our approximate topic-word posterior $q(\phi)$. 

For our methods, we explore several values of sparsity-level $L$.
LightLDA and SparseLDA have no tunable sparsity parameters.
SVIGibbs allows specifying the number of samples $S$ used to approximate $q(z_d)$. We consider $S=\{5, 10\}$, always discarding half of these samples as burn-in. 
For all methods, we set document-topic smoothing $\alpha = 0.5$
and topic-word smoothing $\prior{\lambda} = 0.1$.
We set the stochastic learning rate at iteration $t$ to $\rho_t = (\delta + t)^{-\kappa}$. We use grid search to find the best heldout score on validation data, considering delay $\delta \in \{1, 10\}$ and decay $\kappa \in \{0.55, 0.65\}$.

\begin{figure}[!t]
\setlength{\tabcolsep}{.01cm}
\begin{tabular}{l l}
\begin{minipage}{.75\textwidth}
\includegraphics[height=3.65cm]{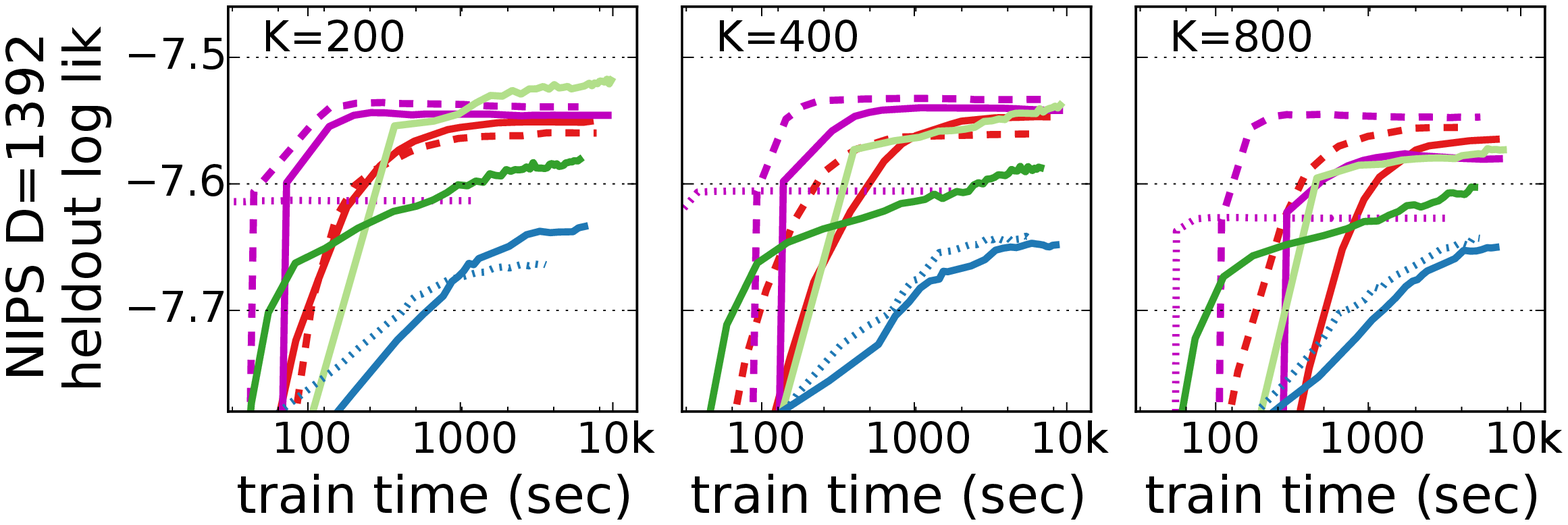}
\\
\includegraphics[height=3.65cm]{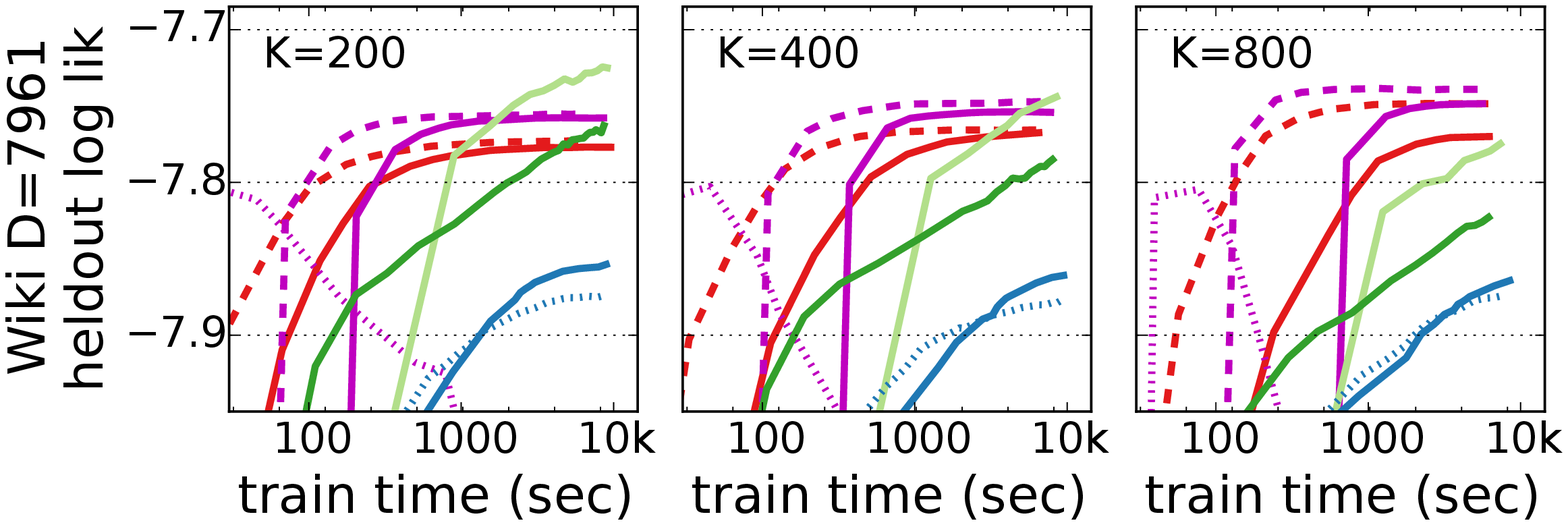}
\end{minipage}
&
\hspace*{-\parindent}
\begin{minipage}{.2\textwidth}
\begin{flushleft}
\includegraphics[
    trim=.5cm 0cm 0cm 0cm,clip=true,
    height=4.2cm
    ]
{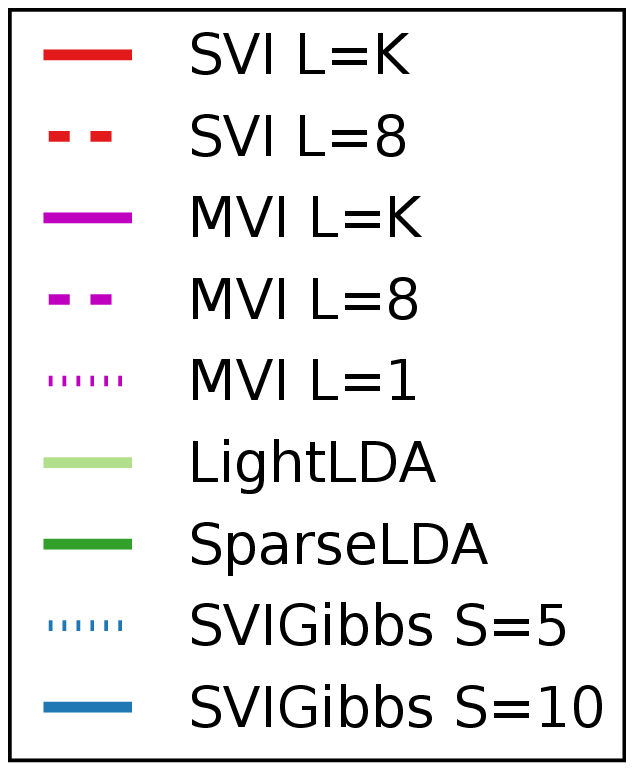}
\end{flushleft}
\end{minipage}
\end{tabular}
\\
\includegraphics[height=3.5cm]{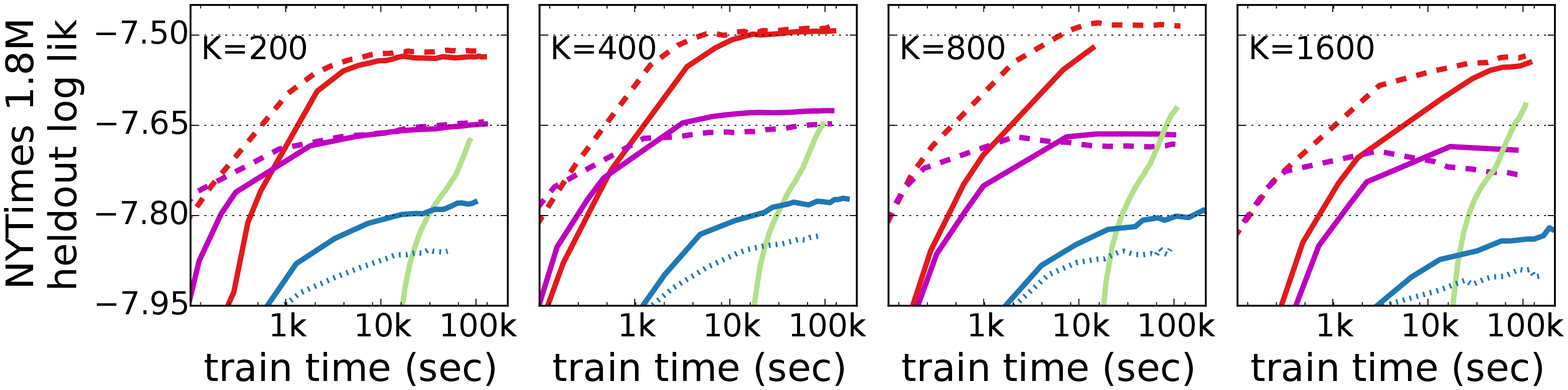}
\caption{
Analysis of 1392 NIPS articles (\emph{top row}), 7961 Wikipedia articles (\emph{middle}), 
and 1.8 million New York Times articles (\emph{bottom}).
We use 200 batches for NY Times and 5 batches otherwise.
Each panel shows for a single $K$ value how heldout likelihood (higher is better) changes over time for sparse and dense versions of our algorithms and external baselines.
Training time is plotted on log-scale.
}
\label{fig:results-NIPS}
\end{figure}

Fig.~\ref{fig:results-NIPS} compares these methods on 3 datasets: 1392 NIPS articles, 7961 Wikipedia articles, and 1.8 million New York Times articles. Each curve represents the best of many random initializations.  
Following \cite{wang:ohdp}, 
we evaluate via heldout likelihoods via a document completion task. 
Given a test document $x_d$, we divide its words at random by type into two pieces: 80\% in $x^{A}_d$ and 20\% in $x^{B}_d$. We use set A to estimate document-topic probabilities $\hat{\pi}_d$, and then evaluate this estimate on set B by computing $\log p( x^B_d | \hat{\pi}_d, \hat{\phi})$.
See supplement for details.
Across all datasets, our conclusions are:

\paragraph{Moderate sparsity tends to be best.}
Throughout Fig.~\ref{fig:results-NIPS}, we see that runs 
with sparsity-level $L=8$ under both memoized and stochastic
algorithms converge several times faster than $L=K$, but yield indistinguishable predictions.
For example, on Wikipedia with $K=800$ both MVI and SVI plateau after 200 seconds with $L=8$, but require over 1000 seconds for best performance with $L=K$.

\paragraph{Hard assignments can fail catastrophically.}
We suspect that $L=1$ is too coarse to accurately capture multiple senses of vocabulary words, instead favoring poor local optima where each word is attracted to a single best topic without regard for other words in the document.
In practice, $L=1$ may either plateau early at noticeably worse performance (e.g., NIPS) or fall into progressively worse local optima (e.g., Wiki).
This failure mode can occur because MVI and SVI for topic models both re-estimate $q(z_d)$ and $q(\pi_d)$ from scratch each time we visit a document.

\paragraph{Baselines converge slowly.}
Throughout Fig.~\ref{fig:results-NIPS}, few runs of SparseLDA or SVIGibbs reaches competitive predictions in the allowed time limit (3 hours for NIPS and Wiki, 2 days for NYTimes).
SVIGibbs benefits from using $S=10$ instead of $S=5$ samples only on NYTimes. More than 10 samples did not improve performance further.
As expected, LightLDA has higher raw throughput than our $L-$sparse MVI or SVI methods, and for small datasets eventually makes slightly better predictions when $K=200$.
However, across all $K$ values we find our $L-$sparse methods reach competitive values faster, especially on the large NYTimes dataset. For large $K$ we find LightLDA never catches up in the allotted time.
Note that LightLDA's speed comes from a Metropolis-Hastings proposal that is highly specialized to topic models of discrete data, while other methods (including our own) are broadly applicable to cluster-based models with non-multinomial likelihoods.

\section{Conclusion}
We have introduced a simple sparsity constraint for approximate posteriors which enjoys faster training times, equal or better heldout predictions, and intuitive interpretation.
Our algorithms can be dropped-in to any ML, MAP, or full-posterior variational clustering objective and 
are easy to parallelize across minibatches.
Unlike previous efforts encouraging sparsity such as Sparse EM \citep{neal:mem} or Expectation Truncation \citep{lucke2010expectation}, we have procedures that easily scale to millions of examples without prohibitive storage costs, we present proof that our chosen top$-L$ selection procedure is optimal, and we have done rigorous experiments demonstrating that often modest values of $L=4$ or $L=8$ are much better than $L=1$.

We have released Python code with fast C++ subroutines to encourage reuse by practioners.
We anticipate further research in adapting $L>1$ sparsity to sequential models like HMMs, to structured variational approximations, to Bayesian nonparametric models with adaptive truncations \citep{hughes:moVB}, and to fast methods like KD-trees for computing cluster weights \citep{moore1999very}.

\newpage

{\small
\bibliography{journal_names.bib,references.bib}
}

\newpage
\begin{appendix}

\section{Mean-field variational for the mixture model}
\label{supp:VariationalForMixtures}

\subsection{Generative model}
Global parameters:
\begin{align}
p(\pi) &= \ts \mbox{Dir}_K( \pi | \frac{\alpha}{K} \ldots \frac{\alpha}{K} )
\\
p(\phi) &= \ts \prod_{k=1}^K 
\mbox{P}( \phi_k | \prior{\lambda} )
\end{align}
where $P$ is a conjugate prior density in the exponential family.

Local assignments $z$ and observed data $x$:
\begin{align}
p(z | \pi) &= \ts \prod_{n=1}^N \mbox{Cat}_K( z_n | \pi_1, \ldots \pi_K)
\\
p(x | z, \phi) &= \ts \prod_{n=1}^N \mbox{F}(x_n | \phi_{z_n})
\end{align}
where $F$ is any likelihood density in the exponential family, with conjugate prior $P$.

\subsection{Assumed mean-field approximate posterior}
Approximate posteriors for global parameters:
\begin{align}
q(\pi | \hat{\theta}) &=
\ts \mbox{Dir}_K( \htheta_{1} \ldots \htheta_{K} )
\\
q(\phi | \hat{\lambda}) &=
\ts \prod_{k=1}^K 
\mbox{P}( \phi_k | \hat{\lambda}_k)
\end{align}
Approximate posterior for local assignment:
\begin{align}
q(z | \hr) &= \ts \prod_{n=1}^{N}
\mbox{Cat}_K(z_{n} | \hr_{n1}, \ldots \hr_{nK}),
\end{align}

\subsection{Evidence lower-bound objective function}
\begin{align}
\label{eq:MixtureModelELBO}
\mathcal{L}(x, \hr, \htheta, \hat{\lambda})
&= \log p(x | \alpha, \prior{\lambda}) - \mbox{KL}(q || p)
\\ \notag
&= \E_q[ \log p(x, z, \pi, \phi) - \log q(z, \pi, \phi) ]
\\ \notag
&= \Ldata(x, \hr, \prior{\lambda}) + \Lentropy(\hr) + \Lalloc(\hr, \htheta)
\end{align}
where we have defined several iterpretable terms which separate the influence of the different free variational parameters.
\begin{align}
\Lalloc(\hr, \htheta) 
&\triangleq \E_{q(\pi | \htheta)q(z|\hr)} [ \log p(z) + \log \frac{p(\pi)}{q(\pi)} ]
\\
\Lentropy(\hr) &\triangleq - \E_{q(z|\hr)}[ \log q(z) ]
\\
\Ldata(x, \hr, \hat{\lambda}) &\triangleq
\E_{
q(z|\hr)
q(\phi|\hat{\lambda})
} [ \log p(x | z, \phi) + \log \frac{p(\phi)}{q(\phi)} ]
\end{align}

\paragraph{Mixture allocation term.} For the mixture model, we can expand the expectation defining $\Lalloc$ and simplify for the following closed-form function:
\begin{align}
\Lalloc(\hr, \htheta) &= \ts
\cDir([\frac{\alpha}{K} ~ \ldots \frac{\alpha}{K}])
- \cDir([\htheta_{1} \ldots \htheta_{K}])
\\ \notag
&\quad + \sum_{k=1}^K \ts
(N_k(\hr) + \frac{\alpha}{K} - \htheta_k )
\Big[
\psi(\htheta_k) - \ts \psi(\sum_{\ell=1}^K \htheta_\ell)
\Big]
\end{align}
where $\psi(\cdot)$ is the digamma function and  $\cDir(\cdot)$ is the log cumulant function, also called the log normalization constant, of the Dirichlet distribution:
\begin{align}
\cDir([a_1, \ldots a_K]) &\triangleq \ts
\log \Gamma(\sum_{k=1}^K a_k) 
- \sum_{k=1}^K \log \Gamma(a_k).
\end{align}
\paragraph{Entropy term.} The entropy of the approximate posterior for cluster assignments is:
\begin{align}
\Lentropy(\hr) &= \ts
- \sum_{n=1}^N \sum_{k=1}^K \hr_{nk} \log \hr_{nk}
\end{align}

\paragraph{Data term.} Evaluating the data term $\Ldata$ requires a particular choice for the likelihood $\mbox{F}$ and prior density $\mbox{P}$. We discuss several cases in Sec.~\ref{supp:Ldata}

\section{Variational methods for data generated by the exponential family}
\label{supp:Ldata}

\subsection{Zero-mean Gaussian likelihood and Wishart prior}

\paragraph{Zero-mean Gaussian likelihood.}
Each observed data vector $x_n$ is a real vector of size $D$. We assume each cluster $k$ has a precision matrix parameter $\Phi_k$ which is symmetric and positive definite. The log likelihood of each observation is then:
\begin{align}
\log F(x_n | \Phi_k)&=
\log \mathcal{N}(x_n | 0, \Phi_k ^{-1})
\\
&= -\frac{D}{2} \log [2\pi]  + \frac{1}{2} \log | \Phi_k | - \frac{1}{2} \mbox{tr}( \Phi_k x_n x_n^T)
\end{align}

\paragraph{Wishart prior.}
The Wishart prior is defined by a positive real $\prior{\nu}$, which can be interpreted as a pseudo-count of prior strength or degrees-of-freedom, and $\prior{\Lambda}$, a $D\times D$ symmetric positive matrix.  The log density of the Wishart prior is given by:
\begin{align}
 \log \mbox{P}( \Phi_k | \prior{\nu}, \prior{\Lambda}) 
 &=
\mbox{c}_{\text{Wish}}(\prior{\nu}, \prior{\Lambda})
+ \frac{\prior{\nu}-D-1}{2} \log |\Phi_k|
-\frac{1}{2} \mbox{tr}(\Phi_k \prior{\Lambda}^{-1})
\end{align}
where the cumulant function is
\begin{align}
c_{\text{Wish}}(\nu, \Lambda) &\triangleq 
 - \frac{\nu D}{2} \log 2 
 - \log \Gamma_{D}\Big( \frac{\nu}{2} \Big)
 + \frac{\nu}{2} \log \Big| \Lambda^{-1} \Big|
\end{align}

where $\Gamma_D(a)$ is the multivariate Gamma function, defined as $\Gamma_D(a) = \pi^{D(D-1)/4} \prod_{d=1}^D \Gamma(a + \frac{1 - d}{2})$.

\paragraph{Approximate variational posterior}
\begin{align}
q(\Phi | \hat{\nu}, \hat{\Lambda})
&= \prod_{k=1}^K \mbox{P}(\Phi_k | \hat{\nu}_k, \hat{\Lambda}_k)
\end{align}

\paragraph{Evaluating the data objective function.}
First, we define sufficient statistic functions for each cluster $k$:
\begin{align}
N_k(\hr) = \sum_{n=1}^N \hr_{nk} \qquad
S_k(x, \hr) = \sum_{n=1}^N \hr_{nk} x_n x_n^T
\end{align}
Then, we can write the data objective as
\begin{align}
\Ldata(x, \hr, \hat{\nu}, \hat{\Lambda})
&\triangleq
\E_{q} \Bigg[
\log p(x | z, \Phi) + \log \frac{p(\Phi)}{q(\Phi)}
\Bigg]
\\ \notag
&= \sum_{n=1}^N \sum_{k=1}^K 
\E_{q(z)}[ \delta_k(z_n) ] \E_{q(\Phi)}[ \log p(x_n | \Phi_k) ]
+ \sum_{k=1}^K \E_{q(\Phi)}[ \log \frac{p(\Phi_k)}{q(\Phi_k)} ]
\\ \notag
&= - \frac{N D}{2} \log [2\pi] 
+ \sum_{k=1}^K 
c_{\text{Wish}}( \prior{\nu}, \prior{\Lambda}) 
- 
c_{\text{Wish}}( \post{\nu}_k, \post{\Lambda}_k)
\\ \notag
&\quad + \sum_{k=1}^K
(N_k(\hr) + \prior{\nu} - \post{\nu}_k)
\E_{q}[ c_F(\Phi_k) ]
\\ \notag
&\quad + \sum_{k=1}^K
(S_k(x, \hr) + \prior{\Lambda} - \post{\Lambda}_k)
\E_{q}[ \Phi_k ]
\end{align}

\subsection{Multinomial likelihood and Dirichlet prior}

\paragraph{Multinomial likelihood.}
Each observation $x_n \in \{1, \ldots V\}$ indicates a single word in a vocabulary of size $V$.
\begin{align}
\log \mbox{F}(x_n | \phi_k)
&= \sum_{v=1}^V \delta_v(x_n) \log \phi_{kv}
\end{align}
The parameter $\phi_k$ is a non-negative vector of $V$ entries that sums to one.

\paragraph{Dirichlet prior.}
We assume $\phi_k$ has a symmetric Dirichlet prior with positive scalar parameter $\prior{\lambda}$:
\begin{align}
\log \mbox{P}(\phi_k | \prior{\lambda})
&= \cDir([\prior{\lambda} \ldots \prior{\lambda}]) + \sum_{v=1}^V (\prior{\lambda}-1) \log \phi_{kv}
\end{align}

\paragraph{Approximate variational posterior.}
We assume that $q(\phi_k)$ is a Dirichlet distribution with parameter $\hat{\lambda}_k$:
\begin{align}
q(\phi | \hat{\lambda}) = 
\ts \prod_{k=1}^K 
\mbox{Dir}_V( \phi_k | \hat{\lambda}_1, \ldots \hat{\lambda}_V)
\end{align}
\paragraph{Evaluating the data objective function.}

\begin{align}
\mathcal{L}_{data}(x, \hr, \hat{\lambda}) 
&\triangleq  \ts
\E_{q}[
\log p(x | z, \phi) 
+ \log 
\frac{p(\phi | \prior{\lambda} )}{q(\phi | \post{\lambda})}
\\ \notag &= 
\sum_{k=1}^K \cDir(\prior{\lambda}) - \cDir(\post{\lambda}_k)
\\ \notag &~~~ 
+ \sum_{k=1}^K \sum_{v=1}^V 
(S_{kv} + \prior{\lambda} - \post{\lambda}_{kv})
\E[ \log \phi_{kv} ]
\end{align}
where $c_D(\cdot)$ is the log cumulant function of the Dirichlet defined above and $S_{kv}$ counts the total number of words of type $v$ assigned to topic $k$.

\section{Mean-field variational for the LDA topic model}
\label{supp:VariationalForLDA}

\subsection{Observed data}
The LDA topic model is a hierarchical mixture applied to data from $D$ documents, $x_1, \ldots x_D$. 
Let each document $x_d$ consist of observed word tokens 
from a fixed vocabulary of $V$ word types, 
though we could easily build a topic model for observations of any type (real, discrete, etc.).
We represent $x_d$ in two ways:
First, as a dense list of the $N_d$ word tokens in document $d$: $x_d = \{ x_{dn} \}_{n=1}^{N_d}$.
Here token $x_{dn} \in \{1, \ldots V\}$ identifies the type of the $n$-th word. Second, we can use a memory-saving sparse histogram representation: $x_d = \{ v_{du}, c_{du} \}_{u=1}^{U_d}$, where $u$ indexes the set of word types that appear at least once in the document, $v_{du} \in \{1, \ldots V\}$ gives the integer id of word type $u$, and $c_{du} \geq 1$ is the count of word type $v_{du}$ in document~$d$. By definition, $\sum_{u=1}^{U_d} c_{du} = \sum_{n=1}^{N_d} x_{dn} = N_d$. 

\subsection{Generative model}

The Latent Dirichlet Allocation (LDA) topic model
generates a document's observations from a mixture model with common topics $\{\phi\}_{k=1}^K$ but document-specific frequencies $\pi_d$.

The model consists of several latent variables. 

\paragraph{Model for global parameters:}
First, we have global topic-word probabilities $\phi = \{ \phi_k \}_{k=1}^K$. Each $\phi_k$ is a non-negative vector of length $V$ (number of words in the vocabulary) that sums to one, such that $\phi_{kv}$ is the probability of type $v$ under topic $k$.
\begin{align}
p(\phi | \prior{\lambda})
&= \prod_{k=1}^K \Dir_V(\phi_k | \prior{\lambda})
\end{align}

\paragraph{Model for local documents:}
Next, each document $d$ contains two local random variables: a document-specific frequency vector $\pi_d$ and token specific assignments $z_d = \{ z_d \}_{n=1}^{N_d}$. These are generated as follows:
\begin{align}
p( \pi | \alpha)
&= \ts
\prod_{d=1}^D \mbox{Dir}_K( \pi_d | \frac{\alpha}{K} \ldots \frac{\alpha}{K} )
\\
p( z | \pi) &= \ts
\prod_{d=1}^D \prod_{n=1}^{N_d} \mbox{Cat}_K( z_{dn} | \pi_d)
\end{align}
Finally, each observed word token $x_{dn}$ is drawn from its assigned topic-word distribution:
\begin{align}
p( x | z, \phi) &= \ts
\prod_{d=1}^D \prod_{n=1}^{N_d}
\mbox{Cat}_V( x_{dn} | \phi_{z_{dn}})
\end{align}

\subsection{Assumed mean-field approximate posterior}
The goal of posterior inference is to estimate the common topics as well as the frequencies and assignments in any document.
The standard mean-field approximate posterior over these quantities is specified by:
\begin{align}
q(z_{d}) &= \ts \prod_{n=1}^{N_d}
\mbox{Cat}_K(z_{dn} | \hr_{dn1}, \ldots \hr_{dnK}),
\\ \notag
q(\pi_d) &= \mbox{Dir}_K(\pi_d | \htheta_{d1}, \ldots \htheta_{dK}),
\\ \notag
q(\phi) &= \ts \prod_{k=1}^K \mbox{Dir}_V(\phi_k | \post{\lambda}_{k1}, \ldots \post{\lambda}_{kV}).
\end{align}

\subsection{Evidence lower-bound objective function}
Under this factorized approximate posterior, we can again set up a variational optimization objective:
\begin{align}
\label{eq:TopicModelELBO}
\mathcal{L}(x, \hr, \htheta, \hat{\lambda})
&= \log p(x) - \mbox{KL}(q || p)
\\ \notag
&= \E_q[ \log p(x, z, \pi, \phi) - \log q(z, \pi, \phi) ]
\end{align}
Just like the mixture model, we can rewrite the terms in this objective as
\begin{align}
\mathcal{L}(x, \hr, \htheta, \hat{\lambda})
&\triangleq \Ldata(x, \hr)
+ \Lentropy(\hr)
+ \Lalloc(\hr, \htheta)
\\
\Ldata &\triangleq \E_q[ \log p(x | z, \phi) + \log \frac{p(\phi)}{q(\phi)} ]
\\
\Lentropy(\hr)
 &\triangleq - \E_q[ \log q(z) ]
\\
\Lalloc &\triangleq \E_q[ \log p(z | \pi) + \log \frac{p(\pi)}{q(\pi)} ]
\end{align}
\paragraph{Entropy term.}
The entropy of the assignments term is simple to compute:
\begin{align}
\Lentropy = - \sum_{d=1}^D \sum_{n=1}^{N_d} \hr_{dnk} \log \hr_{dnk}
\end{align}
This is needed purely for computing the value of the objective function. No parameter updates require this entropy. However, because tracking the objective is useful for diagnosing performance in our SVI and MVI algorithms, we do compute this entropy at every iteration.

\paragraph{Allocation term.}
After expanding the required expectations and simplifying, the term representing the allocation of topics to documents becomes
\begin{align}
\Lalloc(\hr, \htheta) &=
\sum_{d=1}^D 
\Bigg(
\cDir([\frac{\alpha}{K} \ldots \frac{\alpha}{K}])
-
\cDir([\htheta_{d1} \ldots \htheta_{dK}])
\\
&\quad + \sum_{k=1}^K 
[N_{dk}(\hr_d) + \frac{\alpha}{K} - \htheta_{dk}]
[ \psi(\htheta_{dk}) - \psi(\ts \sum_{\ell=1}^K \htheta_{d\ell}) ] \Bigg)
\end{align}
where we have defined the normalization function $\cDir$ of the Dirichlet distribution as:
\begin{align}
\cDir([a_1 \ldots a_K]) \triangleq \log \Gamma(\sum_{\ell=1}^K a_\ell) - \sum_{\ell=1}^K \log \Gamma(a_\ell)
\end{align}
\paragraph{Data term.} The data term expectations are described in Sec.~\ref{supp:Ldata}. See especially the section on multinomial likelihoods.
\section{Algorithms for Topic Model Local Step via Sparse Responsibilities}
\label{supp:TopicModelLocalStep}

As explained in the main paper, coordinate ascent algorithms for the LDA variational objective require the local step for each document $d$ to be \emph{iterative}, alternating between updating $q(z_d | \hr_d)$ and updating $q(\pi_d | \htheta_d)$ until convergence.
Alg.~\ref{alg:TopicModelLocalStep} in the main paper outlines the exact procedures required by the conventional dense algorithm and our new sparse version, presenting the two methods side-by-side to aid comparison.

\subsection{Details of updates for responsibilities.}

As explained in the main paper, under the usual dense representation, the optimal update for the assignment vector of token $n$ has a closed form like the mixture model, but with document-specific weights which depend on the document-topic pseudocounts $\htheta_d$:
\begin{align}
\label{eq:Update_r_dn_TopicModel_supp}
\hr_{dn} &= \textsc{DenseRespFromWeights}([W_{dn1} \ldots W_{dnk}]),
\\ \notag
W_{dnk}(x_{dn}, \htheta, \hat{\lambda}) &\triangleq \E_q[ \log \pi_{dk} + \log \phi_{kx_{dn}} ],
\\ \notag 
\E_q[\log \pi_{dk}] &\triangleq 
\ts \psi(\htheta_{dk}) 
\ts - \psi(\sum_{\ell=1}^K \htheta_{d\ell}).
\end{align}
We can easily incorporate our sparsity-level constraint to enforce at most $L$ non-zero entries in $\hr_{dn}$. In this case, the optimal $L$-sparse vector $\hr_d$ can still be found via the \textsc{TopLRespFromWeights} procedure from the main paper.

\paragraph{Sharing parameters by word type.}
Naively, tracking the assignments for document $d$ requires explicitly representing a separate $K$-dimensional distribution for each of the $N_d$ tokens.
Howeveir, we can save memory and runtime by recognizing that for a token with word type $v$, the optimal value of Eq.~\eqref{eq:Update_r_dn_TopicModel_supp} will be the same for all tokens in the document with the same type. 
We can thus share parameters with no loss in representational power, requiring $U_d$ separate $K$-dimensional distributions, where $\hr_{dn} \triangleq \hr_{du_{dn}}$.

\subsection{Iterative single-document algorithm for dense responsibilities.}
The procedure \textsc{DenseStepForDoc} in Alg.~\ref{alg:TopicModelLocalStep} provides the complete procedure needed to update $\hr_d, \htheta_d$ to a local optima of $\L$ given the global hyperparameter $\alpha > 0$ and global topic-word approximate posteriors $q(\phi_k | \hat{\lambda}_k)$ for each topic $k$.

Following standard practice for dense assignments, \textsc{DenseStepForDoc} is a block-coordinate ascent algorithm that iteratively loops between updating $\hr_d$ and $\htheta_d$.
When computing the log posterior weights $W_{duk}$, two easy speed-ups are possible: First, we need only evaluate $C_{vk} = \E_q[ \log \phi_{kv}]$ once for each word type $v$ and topic $k$ and reuse the value across iterations. Second, we can directly compute the effective log prior probability $P_{dk} \triangleq \E_q[\log \pi_{dk}]$ during iterations, and instantiate $\htheta$ after the algorithm converges.

To initialize the update cycle for a document, we recommend visiting each token $n$ and 
updating it with initial weight $W'_{dnk} = \E_q[ \log \phi_{k x_{dn}} ]$. This essentially assumes the document-topic frequency vector $\pi_d$ is known to be uniform, which is reasonable. This lets the topic-word likelihoods drive the initial assignments. We then alternate between updates until either a maximum number of iterations is reached (typically 100) or the maximum change of all document-topic counts $N_{dk}$ falls below a threshold (typically 0.05).

Each iteration updates $P_{dk}$ with cost $O(K)$,
and then performs $U_d$ evaluations of \textsc{DenseRespFromWeights}, each with dense cost $O(K)$.
On most datasets, we find these local iterations are by far the dominant computational cost.

\subsection{Iterative single-document algorithm for sparse responsibilities.}
The procedure \textsc{LSparseStepForDoc} in Alg.~\ref{alg:TopicModelLocalStep} provides the complete procedure needed to update $\hr_d, \htheta_d$ to a local optima of $\L$ under the addditional constraint that each token's responsibility vector has at most $L$ non-zero entries, 
As discussed in the main paper, 
throughout this algorithm we combine $L-$sparse representation of the responsibilities with the further assumption that once a topic's mass $N_{dk}$ decays near zero, it will never rise again. With this assumption,
at every iteration we identify the set of active topics (those with non-neglible mass) in the document: $\mathcal{A}_d \triangleq \{k : N_{dk} > \epsilon \}$.
Only these topics will have weight large enough to be chosen in the top $L$ for any token.
Thus, throughout \textsc{TopLRespForDoc} we need only loop over the active set. Each iteration costs $O(|\mathcal{A}_d|)$ instead of $O(K)$.

Discarding topics whose mass within a document drops below $\epsilon$ is justified by previous empirical observations of the so-called ``digamma problem'' described in \cite{mimno2012sparsesvi}: for topics with negligible mass, the expected log probability term becomes vanishingly small. For example, $\psi(\frac{\alpha}{K}) \approx -200$ for $\alpha \approx 0.5$ and $K \approx 100$, and gets smaller as $K$ increases.
 
In practice, after the first few iterations the active set stabilizes and each token's top $L$ topics rarely change while the relative responsibilities continue to improve. In this regime, we can amortize the cost of \textsc{LSparseStepForDoc} by avoiding some selection steps altogether, instead treating the previously determined top $L$ indices for each token as fixed and simply reweighting the responsibility values at those tokens.
We perform selection for the first 5 iterations and then only every 10 iterations, which yields large speedups without loss in solution quality.

\subsection{Initialization and Restart proposals for the local step}
\begin{figure}
\centering
\includegraphics[width=.65\textwidth]{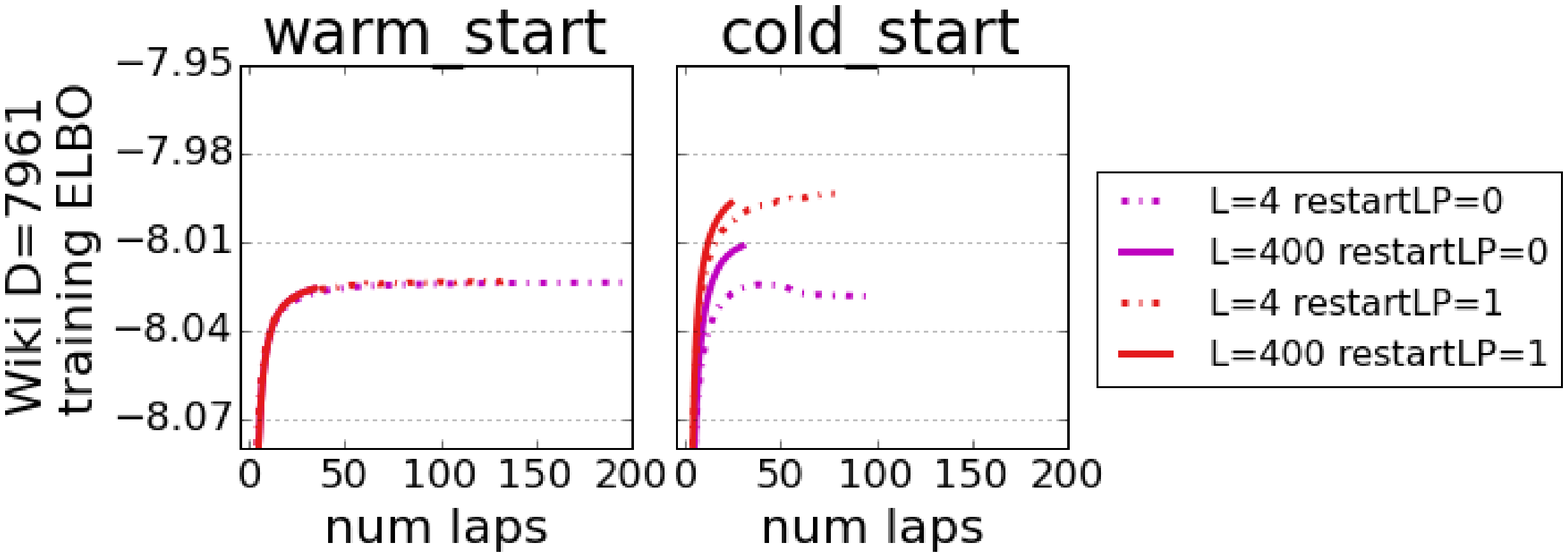}
\\
\includegraphics[width=.65\textwidth]{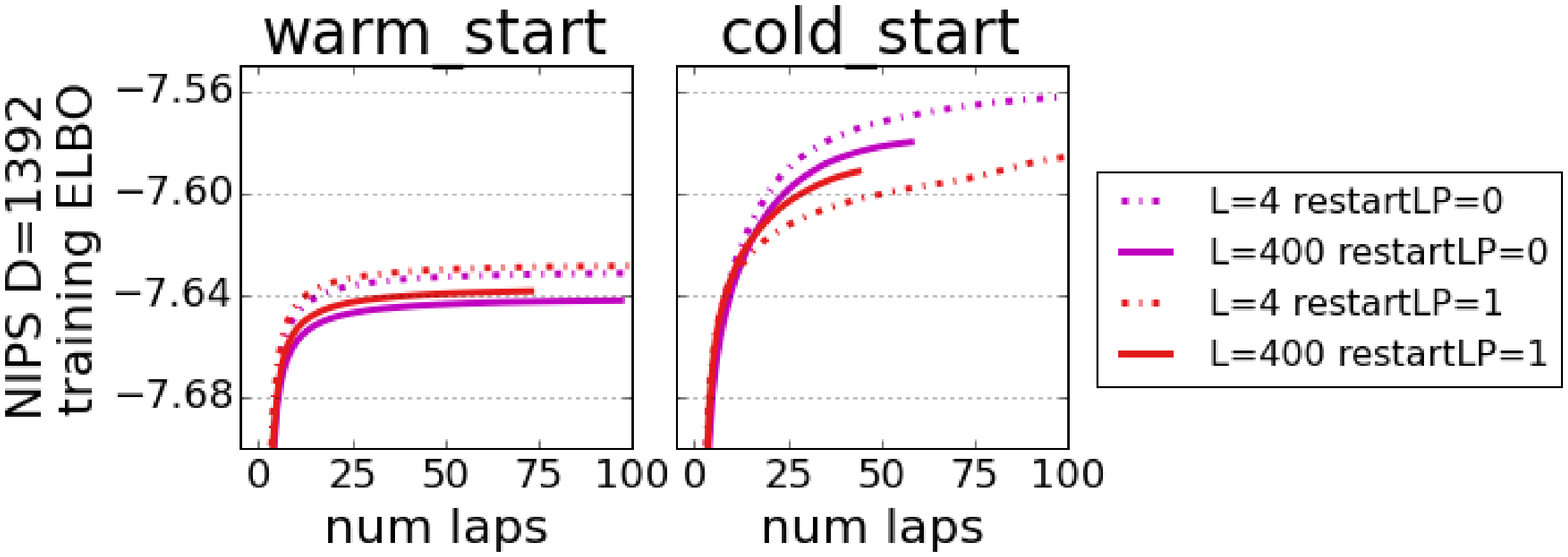}
\\
\includegraphics[width=.65\textwidth]{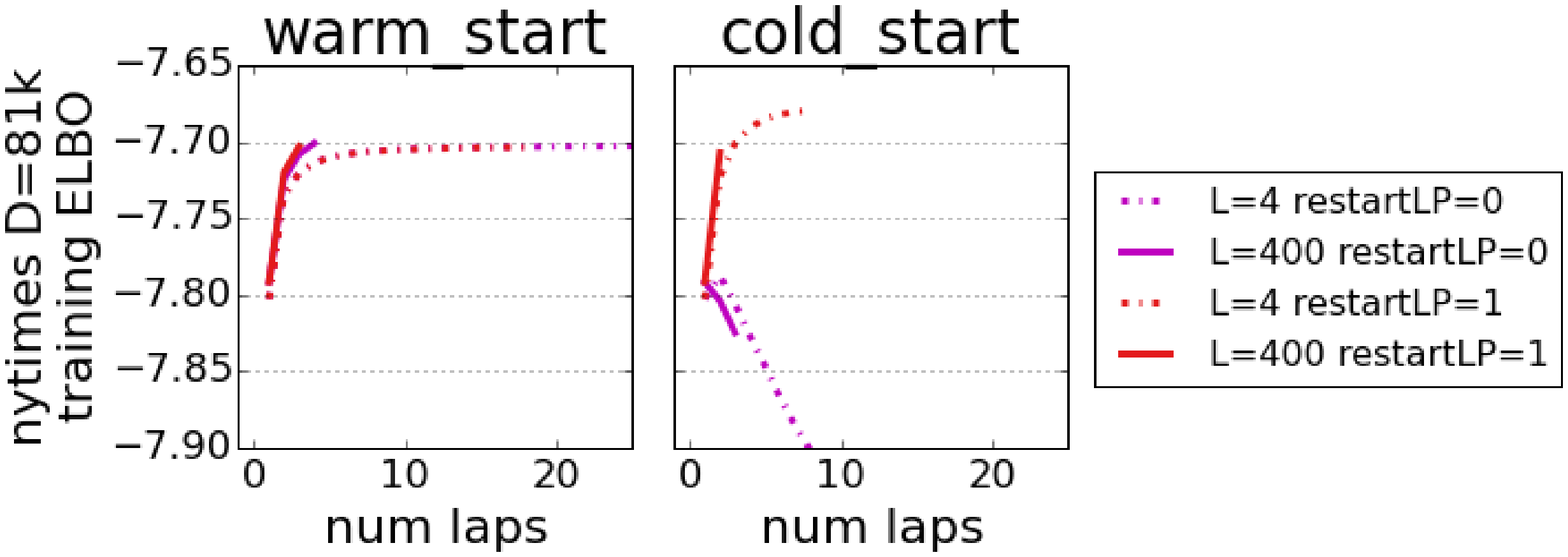}
\caption{
Comparison of warm start (using counts stored from previous visits to each document) and cold start (initializing weights from scratch as recommended in the main paper) for dense and sparse local steps for $K=400$ on a variety of datasets. 
With restart proposals enabled (restartLP=1, red curves), cold start always does as good or better than warm starts.
}
\label{fig:warm_vs_cold}
\end{figure}

In the main paper, we advocate a ``cold start'' strategy for handling repeat visits to a document. This means we do not store any document-specific information, instead initializing weights from scratch as detailed in Alg.~\ref{alg:TopicModelLocalStep}. 
Not only is this more scalable because it avoids storage costs for huge corpuses, but we find that this allows us to reach much higher objective values $\L$ than the alternative ``warm start'' strategy, which would store document-topic counts from previous visits and use these to jumpstart the next local step at each chosen document. Fig.~\ref{fig:warm_vs_cold} shows that while warm starting does allow more complete passes through the dataset (laps) completed, it tends to get stuck in worse local optima. 

The key to making cold start work in practice is using the restart proposals from \citep{hughes:hdpreliable}. Without these, Fig.~\ref{fig:warm_vs_cold} shows that the $\L$ value can decrease badly over time, indicating the inference gets stuck in progressively worse local optima.
However, with restarts enabled (red curves), we find our cold start procedure to be much more reliable.

\section{Heldout likelihood calculation for topic model experiments}
In the main paper's experiments, we evaluate all topic model training algorithms by computing heldout likelihood via a document completion task \citep{wang:ohdp}.
Given a heldout document $x_d$, we divide its words at random by type into two pieces: 80\% in $x^{A}_d$ and 20\% in $x^{B}_d$. We use subset A to estimate the document-specific probabilities $\hat{\pi}_d$, and then evaluate the predictions of this estimate on the remaining words in B.
Throughout, we fix point estimates for each topic $k$
to the trained posterior mean
$\hat{\phi}_k = \E_q[ \phi_k]$. Across many heldout documents, we measure the log-likelihood:
\begin{align}
\mbox{score}(x^A, x^B, \hat{\phi}) 
&= \frac{
\sum_{d} \sum_{n=1}^{|x^B_d|} \log \sum_{k}
\hat{\pi}_{dk} \hat{\phi}_{kx^B_{dn}}
}
{ \sum_{d} | x^B_d | }
\notag
\end{align}
For all algorithms, we fix a point estimate of topics $\hat{\phi}$ from training and then estimate $\hat{\pi}_d$ in the same way: finding the optimal $q(\pi^A_d)$ and $q(z^A_d)$ for the words in the first piece $x^A_d$ by using \textsc{DenseRespForDoc}.
Finally, we take $\hat{\pi}_d = \E_q[ \pi^A_d ]$ and compute the heldout likelihood of $x^B_d$.

\subsection{Dataset statistics}
Our NIPS dataset consists of 1392 training documents, 100 validation, and 248 test documents. The vocabulary size is 13,649.

Our Wikipedia dataset has 7961 training documents, 500 validation, and 500 test documents. The vocabulary size is 6130.

Our NYTimes dataset has 1816800 training documents, 500 validation, and 500 test documents. The vocabulary size is 8000.

\section{Top L Selection Algorithms}
\label{supp:SelectionAlg}

In this section, we discuss how the \textsc{SelectTopL} algorithm introduced in the main paper
would be implemented in practice, since selection algorithms are often unknown to a machine learning audience.
Remember that \textsc{SelectTopL} identifies the top $L$ indices of a provided array of floating-point values.

As part of this supplement, we have released an example code file \texttt{SelectTopL.cpp} whose complete source is in Sec.~\ref{supp:CompleteSourceCode}
. This file offers a simple demo of using selection algorithms to find the top $L$ entries of small, randomly generated vectors.
Below, we first discuss how to execute and interpret the results of this demo program, and then offer a detailed-walk through of the actual code.

\subsection{Using the SelectTopL code}

The provided C++ file called \texttt{SelectTopL.cpp} can be compiled and run using modern C++ compilers, such as the Gnu compiler \texttt{g++}. Our code does require the Eigen library for vectors and matrices, which can be found online at \url{http://eigen.tuxfamily.org}.

\paragraph{Compiling.}
At a standard terminal prompt, we compile the code into an executable.
\begin{verbatim}
g++ -I/path/to/eigen/ -O3 SelectTopL.cpp -o SelectTopLDemo
\end{verbatim}

\paragraph{Running.}
We can then run the executable.
\begin{verbatim}
./SelectTopLDemo
\end{verbatim}

The demo executable will perform several sequential tasks:

\begin{enumerate}
\item Create an unsorted, random weight vector of size $K=10$. Print the indices and values.

\item Sort the vector in descending order, in place. Print the resulting vector's original indices and corresponding values.

\item Call \texttt{SelectTopL}(1), which will place the largest single entry of the vector in the first position. Print the full vector and corresponding indices.

\item Repeat calls to \texttt{SelectTopL}($L$), for each value of $L = \{2, 3, \ldots 9\}$.
\end{enumerate}

\paragraph{Expected output.} The following text will be printed to stdout:

\begin{lstlisting}
seed: 555542
Created random weight vector of size 10
inds:      0     1     2     3     4     5     6     7     8     9
data:   0.35  0.77  0.49  0.41  0.58  0.02  0.26  0.86  0.68  0.16
Calling sortIndices() 
inds:      7     1     8     4     2     3     0     6     9     5
data:   0.86  0.77  0.68  0.58  0.49  0.41  0.35  0.26  0.16  0.02
Reset to original order 
inds:      0     1     2     3     4     5     6     7     8     9
data:   0.35  0.77  0.49  0.41  0.58  0.02  0.26  0.86  0.68  0.16
Calling selectTopLIndices(1)
inds:      7     1     8     4     3     9     6     2     0     5
data:   0.86  0.77  0.68  0.58  0.41  0.16  0.26  0.49  0.35  0.02
Reset to original order 
Calling selectTopLIndices(2)
inds:      7     1     8     4     3     9     6     2     0     5
data:   0.86  0.77  0.68  0.58  0.41  0.16  0.26  0.49  0.35  0.02
Reset to original order 
Calling selectTopLIndices(3)
inds:      7     1     8     4     3     9     6     2     0     5
data:   0.86  0.77  0.68  0.58  0.41  0.16  0.26  0.49  0.35  0.02
Reset to original order 
Calling selectTopLIndices(4)
inds:      7     1     8     4     3     9     6     2     0     5
data:   0.86  0.77  0.68  0.58  0.41  0.16  0.26  0.49  0.35  0.02
Reset to original order 
Calling selectTopLIndices(5)
inds:      8     1     7     4     2     3     0     6     9     5
data:   0.68  0.77  0.86  0.58  0.49  0.41  0.35  0.26  0.16  0.02
Reset to original order 
Calling selectTopLIndices(6)
inds:      8     1     7     4     2     3     0     6     9     5
data:   0.68  0.77  0.86  0.58  0.49  0.41  0.35  0.26  0.16  0.02
Reset to original order 
Calling selectTopLIndices(7)
inds:      8     1     7     4     2     3     0     6     9     5
data:   0.68  0.77  0.86  0.58  0.49  0.41  0.35  0.26  0.16  0.02
Reset to original order 
Calling selectTopLIndices(8)
inds:      8     1     7     4     3     0     2     6     9     5
data:   0.68  0.77  0.86  0.58  0.41  0.35  0.49  0.26  0.16  0.02
Reset to original order 
Calling selectTopLIndices(9)
inds:      8     1     7     4     3     0     2     6     9     5
data:   0.68  0.77  0.86  0.58  0.41  0.35  0.49  0.26  0.16  0.02
Reset to original order 
\end{lstlisting}

\subsection{Remark: Selection is different than sorting}

The \textsc{SelectTopL} procedure is quite different from sorting the array completely and then just returning the top $L$ values. Instead, it uses a recursive algorithm 
whose invariant condition is the following:
given an array with positions $\{0, 1, \ldots K-1\}$, 
guarantee that any value in the first $L$ positions $\{0, 1, \ldots L-1\}$ is larger than any value in the remaining positions $\{L, L+1, \ldots K-1\}$ of the array.

It sometimes happens that the first $L$ values turn out sorted, but there is no guarantee that they will be. For example, in the output above, we see that after calling \texttt{selectTopLIndices(9)}, the first three indices are not strictly in sorted order.

\subsection{Detailed walk-through}

Our implementation defines a simple struct to represent the weight vector data and the corresponding integer indices side-by-side.

\begin{verbatim}
struct ArrayWithIndices {
    double* xptr; // data array
    int* iptr;    // int indices of data array
    int size;     // length of data array
    ...
}
\end{verbatim}

We can construct our struct by providing a pointer to a weight vector of size $K$. The constructor then creates an int array of indices from ${0, 1, \ldots K-1}$.

\begin{verbatim}
    // Constructor
    ArrayWithIndices(double* xptrIN, int sizeIN) {
        xptr = xptrIN;
        size = sizeIN;
        iptr = new int[size];
        fillIndicesInIncreasingOrder(size);
    }
\end{verbatim}

The helper method \texttt{fillIndicesInIncreasingOrder} simply edits the indices array in-place.
\begin{verbatim}
    // Helper method: reset iptr array to 0, 1, ... K-1 
    void fillIndicesInIncreasingOrder(int size) {
        for (int i = 0; i < size; i++) {
            iptr[i] = i;
        }
    }
\end{verbatim}

\paragraph{Sorting indices.}
To understand selection, we can scaffold by first understanding how to sort this struct.
We can sort the indices from largest to smallest by value using the \texttt{sortIndices} method of our struct. This is a thin wrapper around the sort function of the standard library. We provide pointers to the start and end of the region of the array we wish to sort, as well as a custom \emph{comparison} operation, since we want to sort by the values in \texttt{xptr}, rather than \texttt{iptr}.
After executing this method, we are guaranteed that the array region provided is sorted according to the provided comparison.
\begin{verbatim}
    // Sort indices from largest to smallest data value
    void sortIndices() {
        fillIndicesInIncreasingOrder(this->size);
        std::sort(
            this->iptr,
            this->iptr + this->size,
            GreaterThanComparisonByDataValue(this->xptr)
            );
    }
\end{verbatim}
Note that before calling sort, we quickly make sure that the indices are in their default, increasing order. Otherwise, if we called \texttt{sortIndices} twice in a row, we get different results each time because the internal array of indices would be out-of-order the second time.

\paragraph{Custom comparison operator.}
A simple struct defines the custom comparison.
Given two indices $i$ and $j$, we 
return true if the $i$-th element of the data array is larger than the $j$-th element, and false otherwise.
 No memory allocation happens here, we're just passing pointers around.
\begin{verbatim}
struct GreaterThanComparisonByDataValue {
    const double* xptr;

    GreaterThanComparisonByDataValue(const double * xptrIN) {
        xptr = xptrIN;
    }

    bool operator()(int i, int j) {
        return xptr[i] > xptr[j];
    }
};
\end{verbatim}

\paragraph{Selecting the top L indices.}
Just like sorting, our selection algorithm is a simple call to a standard library function: \texttt{nth\_element} \citep{musser1997introselect}. 
This is an introspective selection function
which will rearrange the elements of a provided array region [0, K) in-place.
The function guarantees that for any $i \in [0, L-1]$ in the front region and any $j \in [L, K-1]$ in the remaining region, the $i$-th element of the resulting array will be larger than the $j$-th element.
Again, we provide a custom comparison function so that we can rearrange the indices but make comparisons by the data value of the weights.
\begin{verbatim}
    void selectTopLIndices(int L) {
        assert(L > 0);
        assert(L <= this->size);
        fillIndicesInIncreasingOrder(this->size);
        std::nth_element(
            this->iptr + 0, // region starts at index 0
            this->iptr + L - 1, // partition entries [0,L-1] from [L, end]
            this->iptr + this->size, // region stops at last index
            GreaterThanComparisonByDataValue(this->xptr)
            );
    }
}
\end{verbatim}

\section{Complete source code for SelectTopL.cpp}
\label{supp:CompleteSourceCode}

\begin{lstlisting}
#include <math.h>
#include <assert.h>
#include <stdio.h>
#include <time.h>
#include "Eigen/Dense"

using namespace std;
using namespace Eigen;

// ======================================================== Define types
// ========================================================
// Simple names for array types
typedef Array<double, Dynamic, Dynamic, RowMajor> Arr2D_d;
typedef Array<double, 1, Dynamic, RowMajor> Arr1D_d;
typedef Array<int, 1, Dynamic, RowMajor> Arr1D_i;

// Simple names for array types with externally allocated memory
typedef Map<Arr2D_d> ExtArr2D_d;
typedef Map<Arr1D_d> ExtArr1D_d;
typedef Map<Arr1D_i> ExtArr1D_i;

// ======================================================== Define comparator
struct GreaterThanComparisonByDataValue {
    const double* xptr;

    GreaterThanComparisonByDataValue(const double * xptrIN) {
        xptr = xptrIN;
    }

    bool operator()(int i, int j) {
        return xptr[i] > xptr[j];
    }
};

struct ArrayWithIndices {
    double* xptr; // data array
    int* iptr;    // int indices of data array
    int size;     // total size of data array
    
    // Constructor
    ArrayWithIndices(double* xptrIN, int sizeIN) {
        xptr = xptrIN;
        size = sizeIN;
        iptr = new int[size];
        fillIndicesInIncreasingOrder(size);
    }

    // Helper method: reset iptr array to 0, 1, 2, 3, ... size-1
    void fillIndicesInIncreasingOrder(int size) {
        for (int i = 0; i < size; i++) {
            iptr[i] = i;
        }
    }

    // Helper method: reset iptr array to 0, 1, 2, 3, ... this->size-1
    void fillIndicesInIncreasingOrder() {
        this->fillIndicesInIncreasingOrder(this->size);
    }

    // Sort the indices from largest to smallest data value
    void sortIndices() {
        fillIndicesInIncreasingOrder(this->size);
        std::sort(
            this->iptr,
            this->iptr + this->size,
            GreaterThanComparisonByDataValue(this->xptr)
            );
    }

    // Rearrange indices so first L values are bigger than remaining values
    // No guaranteed order in either the top-L region or remaining region.
    void selectTopLIndices(int L) {
        assert(L > 0);
        assert(L <= this->size);
        fillIndicesInIncreasingOrder(this->size);
        std::nth_element(
            this->iptr + 0, // region starts at index 0
            this->iptr + L - 1, // first L elements of iptr[start:stop] 
            this->iptr + this->size, // region stops at last index
            GreaterThanComparisonByDataValue(this->xptr)
            );
    }

    // Pretty print the current index order and associated data values
    void pprint() {
        this->pprint(this->size);
    }

    // Pretty print the current index order and associated data values
    void pprint(int topL) {
        printf("inds: ");
        for (int i = 0; i < topL; i++) {
            printf(" %5d",
                this->iptr[i]);
        }
        printf("\n");
        printf("data: ");
        for (int i = 0; i < topL; i++) {
            printf(" % 5.2f",
                this->xptr[this->iptr[i]]);
        }
        printf("\n");
    }

};


int main(int argc, char** argv) {
    // Use fixed random seed for reproducability
    unsigned int seed = 555542;
    //unsigned int seed = 1456763546;
    //unsigned int seed = (unsigned int) time(0);

    printf("seed: %d\n", seed);
    std::srand(seed);

    // Set size of our weight vector
    int K = 10;

    // Generate weight vector of values in [0.0, 1.0] 
    Arr1D_d weightVec = Arr1D_d::Random(K); // Values in [-1, +1]
    weightVec += 1.0;
    weightVec /= 2.0;

    // Create struct with corresponding indices
    ArrayWithIndices myDataAndInds = ArrayWithIndices(
        weightVec.data(), K);
    printf("Created random weight vector of size %d\n", K);
    myDataAndInds.pprint();

    printf("Calling sortIndices() \n");
    myDataAndInds.sortIndices();
    myDataAndInds.pprint();

    printf("Reset to original order \n");
    myDataAndInds.fillIndicesInIncreasingOrder();
    myDataAndInds.pprint();

    for (int L = 1; L < 10; L++) {
        printf("Calling selectTopLIndices(%d)\n", L);

        myDataAndInds.selectTopLIndices(L);
        myDataAndInds.pprint();

        printf("Reset to original order \n");
        myDataAndInds.fillIndicesInIncreasingOrder();
        //myDataAndInds.pprint();
    }

    return 0;
}
\end{lstlisting}

\end{appendix}

\end{document}